\documentclass [11pt]{article}
\usepackage{geometry}                		
\usepackage{graphicx}				
\usepackage{amsmath}
\usepackage{amssymb}

\usepackage{hyperref}
\usepackage{subfig}   
\usepackage{soul}
\title{Hierarchical Predictive Coding Models in a Deep-Learning Framework}
\author{Anthony S. Maida\\Matin Hosseini}

\begin{document}
\maketitle

\begin{abstract}
Bayesian predictive coding is a putative neuromorphic method for acquiring higher-level neural representations
to account for sensory input.
Although originating in the neuroscience community,
there are also efforts in the machine learning community to study these models.
This paper reviews some of the more well known models.
Our review analyzes module connectivity and patterns of information transfer,
seeking to find general principles used across the models.
We also survey some recent attempts to cast these models within a deep learning framework.

A defining feature of Bayesian predictive coding is that it uses top-down, reconstructive mechanisms to predict incoming sensory
inputs or their lower-level representations.
Discrepancies between the predicted and the actual inputs, known as prediction errors,
then give rise to future learning that refines and improves the predictive accuracy of learned higher-level representations.

Predictive coding models intended to describe computations in the neocortex emerged prior to 
the development of deep learning 
and used a
communication structure between modules that we name the Rao-Ballard protocol.
This protocol was derived from a Bayesian generative model with some rather strong statistical
assumptions.
The RB protocol provides a rubric to assess the fidelity of deep learning models
that claim to implement predictive coding.
Here,
we consider the implementation of this protocol within convenient deep learning frameworks
that are robust to less restrictive statistical assumptions.
We employ deep learning examples to
demonstrate the feasibility and effectiveness of predictive learning.
Although neurally inspired in a historical sense, modern deep learning frameworks can be used at
an intermediate modeling level that is agnostic to specific claims about biologically based neural representations.
The intent is to make the classical models easier to study and more versatile and comprehensive while retaining
their core properties.
\\

\end{abstract}
%
\section{Introduction}
Scientists have hypothesized that the brain, specifically the neocortex, acquires and updates internal representations
by invoking learning mechanisms designed to predict incoming sensory
input.
These predictions are compared with actual sensory input to calculate discrepancies
between input and predictions.
These discrepancies,
known as prediction errors,
\cite{Mumford1992,rao1999predictive,Friston2005,Friston2018a} guide learning in the higher layers to reduce future
prediction errors.

Although many forms of predictive coding have been proposed and found to exist in the brain 
\cite{Aitchison2017}, in regard to the neocortex,
there are two main approaches toward building neural architectures for hierarchical predictive coding.
The first,
exemplified by the references cited above, uses a generative Bayesian approach,
and is called
\textit{Bayesian predictive coding} \cite{Aitchison2017}.
In this approach,
the higher layers learn to reconstruct the outputs of the immediately lower layers.
Thus the higher layers, although represented more abstractly, contain the same information
represented in the lower layers.
The second approach,
exemplified by the work of \cite{Lotter2017}
and \cite{Alexander2015,Alexander2018},
uses the higher layers to \textit{predict the prediction error} in the immediately lower layer.
This knowledge about the prediction error is used to modify the lower layer representation to reduce
future prediction errors.
The acquired representations are presumed to learn spatially and temporally immediate and distant 
causes or modulations of
the input and this gives the representations their ability to predict future inputs.

Predictive coding models inspired by neocortex invariably use feedback from higher layers to
lower layers.
State-of-the-art deep learning models found in applied artificial intelligence
rarely use feedback connections in their calculations.
Both fully connected and convolutional deep networks are purely feedforward (what about semantic
segmentation?).
Recurrent networks, including gated recurrent networks, have lateral connections but not
feedback connections \cite{Goodfellow2016} but not feedback connections across layers.
The models in the deep learning literature which do have feedback connections are actually types
of predictive coding models \cite{Lotter2017,Wen2018a,Han2018a}.

Feedback connections are ubiquitous in the neocortex.
Virtually every feedforward connection from one cortical area to another is accompanied
by a separate feedback connections linking those areas \cite{Felleman1991}.
Both the feedforward connections and feedback connections consume significant brain volume
and would not exist without an important functional reason.
Predictive coding models, both 
Bayesian and non-Bayesian, offer attractive hypotheses for shedding light on the 
the computational purpose of feedback connections,
as well as potentially offering significant improvements in deep learning performance.

As already mentioned,
this type of learning has been called predictive coding,
or more recently predictive processing \cite{Clark2016, Keller2018a}, 
and is usually described using a Bayesian generative model \cite{Lee2003}
where sensory input is combined with prior expectations, to make better future predictions.
The term ``predictive coding''
predates Bayesian predictive coding.
Non-Bayesian predictive coding has been used for other functions such as
reducing information transmission requirements and cancelling the effects of
self-generated actions \cite{Aitchison2017}.

Predictive coding can be usefully viewed as a type of representation learning \cite{Bengio2014a}.
The learning mechanisms to support Bayesian predictive coding should
improve the quality of internal representations as a
side effect of reducing future prediction errors.
The predictive/reconstructive capacity ensures that the acquired representations can
fully represent information contained in the input.
Since learning is driven by prediction errors, the learning is unsupervised and simply
requires a data stream of information to be predicted.
\cite{Bengio2014a} has noted that good representations are effective because
they, in part, capture 
underlying ``explanatory factors of variation behind the data'',
and Bayesian predictive coding shares this goal.

In regard to applications,
predictive coding networks can be used for learning overlapping image components \cite{Spratling2009a},
object classification \cite{Spratling2017b,Wen2018a,Han2018a},
video prediction \cite{Lotter2017},
video anomaly detection \cite{wopauli2019a},
modeling biased competition \cite{Spratling2008a},
response properties of primate visual cortex \cite{rao1999predictive,Spratling2010a},
and EEG evoked brain responses \cite{Friston2005} health concerns\cite{zolnoorimining,goudarzvand2019early}.
It has also been proposed as a unified theory of neocortical function 
\cite{Friston2010,Spratling2017a,Keller2018a}.

Its use in deep learning to build large models has so far been limited,
with the models of \cite{Lotter2017}, \cite{Wen2018a}, and \cite{Han2018a} being perhaps the only examples.
Most predictive coding models have been implemented prior to the emergence of deep learning
frameworks \cite{rao1999predictive,Friston2005,Spratling2009a},
so these classical models are small and do not use 
advanced modules for temporal processing, such as gated recurrent networks \cite{Greff2017}, 
now found
in deep learning frameworks.

This survey explains some early predictive coding models and considers the question
of enhancing them by  using modern deep learning tools.
We analyze the reasons for their effectiveness and offer proposals for future research.
We also introduce a template for module connectivity which we call the Rao-Ballard protocol.
We suggest that it can be used as a rubric to assess the fidelity of deep learning models that claim to implement
predictive coding.
\section{Hierarchical generative models and prediction error}
\subsection{Historical Motivation for Predictive Coding}
An early motivation of predictive coding was to create models of the retina
that had reduced information transfer requirements \cite{Spratling2017a}
because the optic nerve is a bottleneck in transferring visual information to the brain.
Continuous visual input has high redundancy.
If the cells on the retina calculated a moving average of the incoming light, 
both spatially and temporally, this would yield a prediction of the current input.
By comparing the actual input with the prediction, the retina could send a prediction
error to later processing areas, thereby reducing transmission bandwidth.

If the vertebrate retina created a representation of the current input image from scratch,
the information transfer requirements would depend on the bandwidth needed
to build a full representation of the current environment.
However, the brain can generally make a very good prediction of the current state
of the environment using its existing representations of the past environment
along with constraints about how
the environment tends to change.
The prediction error might require much less bandwidth to represent than
building a representation of the current envronmental state from scratch.
If the brain already had a fairly accurate default prediction, then updating the
default prediction with the prediction error could create an
up-to-date representation with lower
information transfer demands.
Furthermore, it is plausible that prediction errors could provide good information
to guide learning to improve the representation and reduce future prediction
errors.

\subsection{Bayesian Predictive Coding}
A more recent attraction of predictive coding models is that they can be formulated
in a
generative \cite{Friston2018a} fashion, enabling them to
predict sensory input.
This is usually expressed as a hierarchy where a higher layer predicts the
outputs of a lower layer (i.e., input to the higher layer)
and any prediction errors provide information to guide
learning in the higher layer.
A hierarchical architecture
involving feedforward and feedback connections
is also consistent with the architecture of the primate neocortex \cite{Felleman1991}.
In the context of perceptual inference and recognition,
a trained generative model has the property that the learned representation can reconstruct
the distribution of raw sensory inputs and estimate hidden causes in the input at
different spatial and temporal scales.
The reconstruction is possible because the learned representation(s) capture
the causal factors that generated the input in the first place.
It requires an inverse mapping from sensations to causes in order to construct
the representation.
This is challenging because computing the inverse mapping 
between sensations and
causes in the physical world is an ill-posed problem, meaning that its
solution is not unique.
In a Bayesian approach, this can be addressed by using appropriate prior probabilities,
which in some cases can be learned if given sufficient input.
One way to do this is by learning to minimize prediction errors
(i.e., predictive coding).

Early models of predictive coding were expressed as statistical models
such as the hierarchical expectation maximization and 
variational free energy models used in \cite{Friston2005}.
Although these models presented an exact calculus for calculating predictions,
the models presented challenges for further study because implementing the calculations
was computationally intensive and the assumptions of the models were highly restrictive.
In very recent work,
the predictive coding idea has entered the realm of deep neural networks in the work of 
\cite{Chalasani2013a,Wen2018a}, and \cite{Lotter2017}.
This is significant because the large toolset that comes with deep learning frameworks
can be employed in an off-the-shelf manner. 
Although such models sacrifice the exact Bayesian calculus,
they can be specified as a deep neural network rather than an abstract statistical model.
This allows the construction of more complex models.
For example, the
\cite{Lotter2017} model known as PredNet, links predictive error modules with convolutional LSTMs \cite{Shi2015a}
and then uses a hierarchy of predictive error modules for next-frame video prediction.
In PredNet,
these components are all specified using the Keras deep learning framework
with a few hundred lines of code.

Some ideas behind predictive coding resemble other types of deep learning
models such as auto-encoders and deep belief networks.
Auto-encoders are explicitly trained to learn representations that can reconstruct their
input.
Deep belief networks are built by stacking restricted Bolztmann machines (RBMs).
RBMs in turn are trained to reconstruct their input.
Comparison of these methods may uncover more universal principles
but are beyond the scope of this survey.

\section{Predictive coding to learn hierarchical context}
\label{sec_basicAssumptions}

Predictive coding models 
construct representation hierarchies by reducing prediction error
across layers.
There is more than one way to approach this depending on the nature
of the representation that the higher layers are intended to construct.
The first way is to construct increasingly abstract feature hierarchies,
by using larger input context later in the hierarchy,
similar to convolutional networks.
The second way is to learn a hierarchy of higher-order errors as
one would in a Taylor series expansion.
Section~\ref{sec_predictiveElements} and~\ref{section_Friston_model} present examples of the first method.
Section~\ref{sub_LotterModel} presents an example of the second method.

\subsection{The Rao/Ballard model}
\label{sec_predictiveElements}
An early neural network demonstration of the context-based approach
involves constructing a small feature hierarchy that models
end-stopped neural receptive fields in primary visual cortex \cite{rao1999predictive}.
This is a landmark paper, 
is somewhat challenging to understand, and is the standard entry point to the literature
on Bayesian predictive coding networks.

The network has three layers which is the minimum number of layers to exhibit
acquired context
effects.
Each layer has two functionally distinct neural classes. 
These classes are \emph{internal representation neurons} and \emph{prediction error neurons}.
Layer~2 error cells in the 
model simulate cells in primary visual cortex and capture context effects
known
to occur in some of these cells.
Many response properties of 
some cells in primary visual cortex are determined by the structure of a feedforward input field, 
known as the classical receptive
field, and these properties are not context sensitive.
However,
response properties of 
other
cells can be affected by presenting stimuli that
fall outside of the classical input region.
These are called non-classical receptive field properties and are attributed to context effects
assumed to come from either lateral or top-down feedback inputs which are
outside
the feedforward field.
The phenomenon of
end-stopping (i.e., orientation-tuned cell response decreases when 
spatial extent of stimulating input exceeds classical receptive field boundary; also known as ``length suppression'')
is one such effect.
End-stopping is caused by surrounding context that
is outside of the feedforward inputs to the the cell
and cell firing rate, counter-intuitively, decreases when the size of a line or edge-shaped
input grows larger than the width of the classical, feedforward receptive field.
\cite{rao1999predictive} show how this can be accounted for by predictive coding.
The \cite{rao1999predictive} model hypothesizes that this effect comes from feedback
inputs originating in a layer that encompasses larger spatial context,
and experimental evidence supports this \cite{Gilbert1986}.

\begin{figure} \centering
\subfloat[]
{
\includegraphics[scale=0.65]{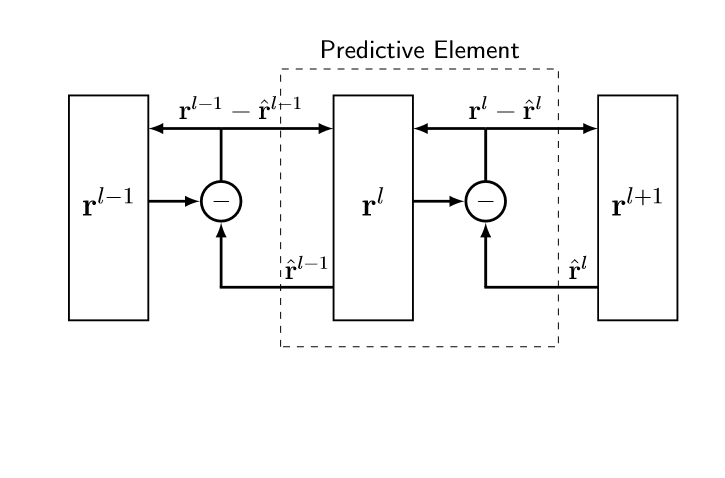}
} \qquad
\subfloat[]
{
\includegraphics[scale=0.65]{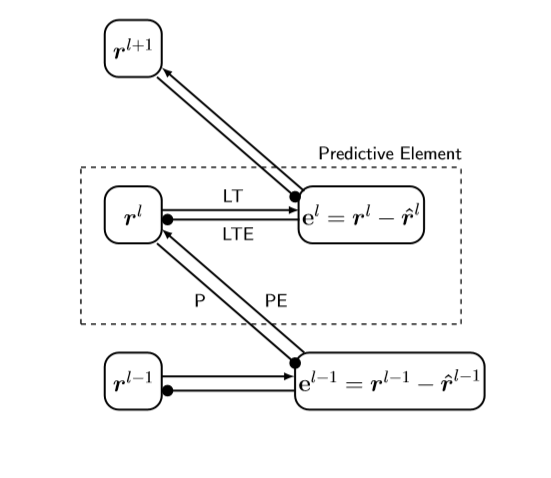}
}
\caption{(a) Rao/Ballard diagram of a predictive element. A predictive element enclosed in the dashed box is
a building block for a predictive coding hierarchy.
A circle surrounding a minus sign denotes a vector of error units
that calculates predictive error, $\mathbf{r}^{l-1}-\hat{\mathbf{r}}^{l-1}$, which is sent both feedforward and
feedback.
(b) 
A data flow diagram that makes
the Rao-Ballard protocol more apparent.
Predictive error and level is made explicit using $\mathbf{e}^l$.
Circle arrowheads indicate subtraction.
The four links associated with a predictive element are now labeled.
(P: prediction, PE: predictive error, LT: lateral target, LTE: lateral target error)}
\label{fig_predictiveElement}
\end{figure}

\cite{rao1999predictive} use a 3-level representation hierarchy where feedback from
a higher layer conveys predictions
of neural activity for the preceding layer.
For instance,
predicted activity in the lowest layer is compared with actual activity 
(raw sensory input). 
Prediction error is the output of a layer and is forwarded to the next higher layer.
Within the hierarchy, there are two classes of neurons: internal representation neurons and 
prediction error neurons.
Weight training is controlled by gradient descent on a cost function based on 
prediction error magnitude.
End-stopping is modeled by letting the highest-level representation use
adjacent spatial context when predicting the middle-level representation.
The highest layer builds a representation for the larger
spatial context.

This model is built around the \textit{predictive element} (PE).
One can view the predictive element as a processing stage or cortical
layer in the brain.
In deep learning terminology, it consists of two neural layers 
performing complementary functions, and connected by both feedforward
and feedback connections.

Predictive elements are stacked into a hierarchy (Fig.~\ref{fig_predictiveElement}(a)).
An element receives prediction errors (via forward connections) from the preceding layer in the hierarchy
and sends predictions in the form of prior probabilities (via backward connections) to the preceding layer.
In Fig.~\ref{fig_predictiveElement}(a), we see the original view of the information flow.
Layer $l+1$ learns a transformed representation of layer $l$
such that its prediction performance for layer $l$'s activity improves.
The representation, $\mathbf{r}^{(l)}$, in a layer represents the
hypothesized causes of the input, but at a higher level of description than the preceding 
layers.
The different layers $l$ provide different representations of the same
causes at increasingly higher description levels.
The representation at each layer is manifested as a set of activation levels
for the vector of neurons forming the layer.

\begin{figure}
\includegraphics[scale=1]{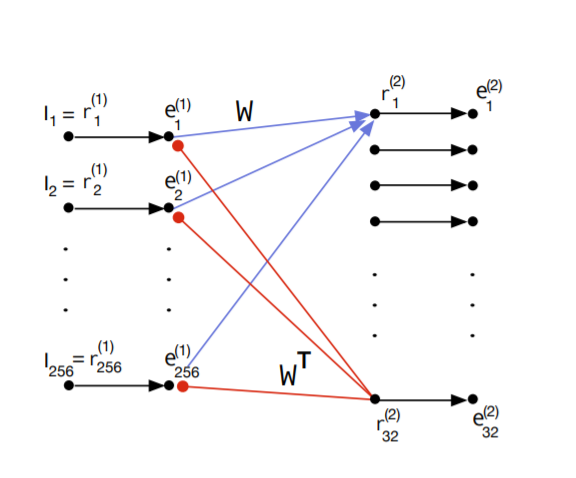}
\caption{Predictive element. Network-level representation of Layer 1 and part of Layer 2 in the \cite{rao1999predictive} model.
$\boldsymbol{e}$ means residual-error unit, $r$ means representation unit, and $I$
means input.
Small black circles represent neurons.
Red arrows ending with solid circles indicate subtractive feedback inhibition.
Red arrows represent the P link, blue arrows represent the PE link, and black arrows
represent the LT link.}
\label{fig_PE_level_1}
\end{figure}

Figure~\ref{fig_predictiveElement}(b) refactors the drawing in part (a) to make the 
predictive error for a layer, $l$, more explicit by using the symbol $\boldsymbol{e}^l$
to represent prediction error
and pairing it with a representation $\boldsymbol{r}^{(l)}$.
This alternative view resembles that presented in 
\cite{Friston2009} and \cite{Friston2010}.
The view in Fig.~\ref{fig_predictiveElement}(b)
shows
that the interaction between adjacent layers obeys a constrained protocol
which we will call the \textit{Rao-Ballard protocol} because of its first implicit appearance in \cite{rao1999predictive}.
In our representation, 
there are four link types: prediction (P), prediction error (PE),
lateral target (LT), and lateral target error (LTE)\@.
Layer output is the information provided by the PE link.
The P and PE links are fully connected and the 
LT and LTE are point-to-point connections (see Fig.~\ref{fig_PE_level_1}).
Notice further that the representation modules only communicate with prediction-error modules
and prediction-error modules only communicate with representation modules.
Prediction error neurons never project downward in the hierarchy and internal representation neurons
never project upward in the hierarchy.
This view makes it easier to  
see the constraints imposed by the Rao-Ballard protocol.
Fig.~\ref{fig_predictiveElement}(b) poses an unanswered question: what kind of hierarchical representations
are acquired in layers $\boldsymbol{r}^{(l-1)}$ through $\boldsymbol{r}^{(l+1)}$ and how do they compare
with more traditional hierarchical representations obtained in deep learning models, 
such as convolutional networks?

Fig.~\ref{fig_PE_level_1} shows a network-level representation of a first-layer 
PE in \cite{rao1999predictive}.
For simplicity, the representation units in Layer 2, $\boldsymbol{r}^{(2)}$, are assumed to be linear.
There are 256 inputs representing the input pixel intensities of a $16 \times 16$ image patch
but only 32 representation elements in the receiving layer, $\boldsymbol{r}^{(2)}$.
The feedforward connections are in blue, $W$, and the feedback connections, $W^T$, are in red.
The top-down prediction, denoted $\boldsymbol{\hat{I}}$, is formed by calculating $W^T\boldsymbol{r}^{(2)}$.
The $\boldsymbol{e}^{(1)}$ units calculate prediction error according to the formula
$\boldsymbol{I} - \boldsymbol{\hat{I}}$.
These ideas are related to the operation of the architecture  and can be written

\begin{subequations}
\label{eq_RB_set}
\begin{align}
\boldsymbol{I} = \boldsymbol{r}^{(1)} \approx \boldsymbol{\hat{I}} &= \mathbf{\hat{r}}^{(1)} = W^T\boldsymbol{r}^{(2)}\label{eq_RB_I}\\
\boldsymbol{e}^{(1)} &=\boldsymbol{I} - \boldsymbol{\hat{I}}\label{eq_RB_predErr}\\
\boldsymbol{r}^{(2)} &\leftarrow \boldsymbol{r}^{(2)} + k_1 W \boldsymbol{e}^{(1)}\label{eq_RB_predictive_update}
\end{align}
\end{subequations}

\noindent
Eqn.~\ref{eq_RB_I} asserts several things.
First, the input $\boldsymbol{I}$ and the first layer representation $\boldsymbol{r}^{(1)}$ are aliases.
$\boldsymbol{I}$ and $\boldsymbol{\hat{I}}$ both have 
dimensions $256 \times1$ (assuming a $16\times 16$ input image patch).
The predicted input $\boldsymbol{\hat{I}}$ and the predicted representation $\boldsymbol{\hat{r}}^{(1)}$
are also aliases.
The predicted input is given by $W^T \boldsymbol{r}^{(2)}$,
where $\boldsymbol{r}^{(2)}$ has dimension $32 \times 1$ and $W$ has dimension
$32\times 256$.
Finally, under normal operating conditions, the input $\boldsymbol{I}$ and
the predicted input $\boldsymbol{\hat{I}}$ should be approximately equal.

Eqn.~\ref{eq_RB_predErr} simply defines the predictive error, $\boldsymbol{e}^{(1)}$, for Layer~1, as
the elementwise difference between the actual and predicted input.
Eqn.~\ref{eq_RB_predictive_update} is an equation for updating the internal representation
in Layer 2 as a function of prediction error.
We can derive Eqn.~\ref{eq_RB_predictive_update} as follows \cite{rao1997predictive}.
Start with a cost function, $J$, on the sum-of-squared predictive error.

\begin{equation}
\label{eq_RB_ssquaresLoss}
J = || \boldsymbol{e} ||^2 = \left(\boldsymbol{I} - W^T\boldsymbol{r} \right)^T \left(\boldsymbol{I} - W^T\boldsymbol{r}\right)
\end{equation}

\noindent
Since the equations are for a one-layer network,
we leave out the layer superscripts for readability.
This cost function does not take into account priors although the 
cost function reported in the original \cite{rao1999predictive} model does incorporate priors.
To prepare for gradient descent,
we obtain the derivative of $J$ with respect to $\boldsymbol{r}$.

\begin{subequations}
\label{eq_RB_deriveOfJ}
\begin{align}
\frac{\partial J}{\partial \boldsymbol{r}} = \frac{\partial \left( \boldsymbol{I} - W^T\boldsymbol{r} \right)^T \left(\boldsymbol{I} - W^T\boldsymbol{r}\right)}{\partial \boldsymbol{r}} &=\\
\left[\frac{\partial \boldsymbol{I} - W^T \boldsymbol{r}}{\partial \boldsymbol{r}}\right]^T 
\frac{\partial \left(\boldsymbol{I} - W^T \boldsymbol{r}\right)^T \left(\boldsymbol{I} - W^T \boldsymbol{r}\right)}{\partial \boldsymbol{I} - W^T \boldsymbol{r}} &= -2 W \boldsymbol{e}
\end{align}
\end{subequations}

\noindent
The intermediate step above uses the chain rule for vectors.
For gradient descent, we travel in the opposite direction of the
derivative at some rate, $k_1 \ge 0$.
This is shown below, were the update time steps, $t$, are made explicit.

\begin{equation}
\label{eq_RB_predUpdateWithT}
\boldsymbol{r}(t+1) = \boldsymbol{r}(t) + k_1 W \boldsymbol{e}(t)
\end{equation}

\noindent
A more compact and convenient way of representing the above formula
is shown in Eqn.~\ref{eq_RB_predictive_update}.

%
%
%

By taking the derivative of $W^T$ with respect to $J$,
an equation for learning in the network can be obtained is shown below.

\begin{equation}
\label{eq_RB_predLearningWithT}
W^T \left(t+1\right) = W^T (t) + k_2 \boldsymbol{e}(t) \boldsymbol{r}^T(t)
\end{equation}

Not shown in Figs.~\ref{fig_predictiveElement} and~\ref{fig_PE_level_1} is the option for higher layers to receive a wider range of inputs,
either from adjacent areas of an input image, from complementary sensory modalities,
or from a longer time scale.
In these cases, the predictions sent back from the higher layer to the previous layer
are influenced by broader context.

\begin{figure}
\includegraphics[scale=0.7]{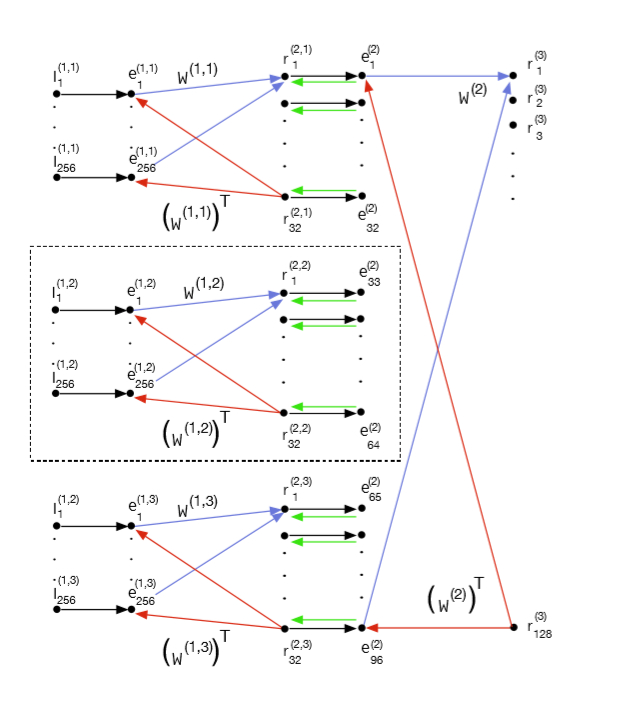}
\caption{Global structure of the \cite{rao1999predictive} model showing hierarchy and adjacent context.
The network from Fig.~\ref{fig_PE_level_1} is inside the dashed box.}
\label{fig_PE_level_global}
\end{figure}

\begin{figure}
\includegraphics[viewport=0.8in 8.7in 8.0in 9.5in,clip=true,scale=0.9]{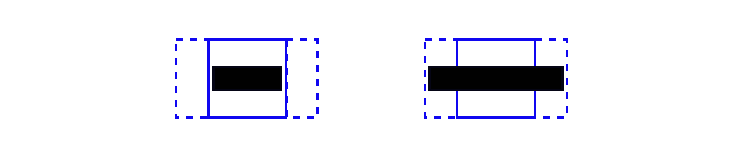}
\caption{Set up for the end-stopping effect.
Three overlapping receptive fields with a bar stimulus.
Error neurons $e_{33}^{(2)}$ -- $e_{64}^{(2)}$ in Fig.~\ref{fig_PE_level_global},
which come from the center receptive field, 
reduce their activation when the bar stimulus on the left is replaced
by the one on the right.}
\label{fig_barStimuli_RaoBallard}
\end{figure}

Given the module in Fig.~\ref{fig_PE_level_1},
we want to see how the larger architecture looks and what happens when it is embedded within a context hierarchy,
in our case, immediately adjacent spatial context.
To achieve this,
Fig.~\ref{fig_PE_level_global} extends Fig.~\ref{fig_PE_level_1} with 
two lateral PEs in the first layer (flanking the original patch module) and an added PE in the second layer.
The input to Layer 2 consists of three overlapping $16\times 16$ image patches 
(patch overlap is 11 pixels,
yielding a global input patch of $16\times 26$).
Fig.~\ref{fig_barStimuli_RaoBallard} shows the overlapping image patches with a superimposed bar stimulus
of two different sizes.
The square with the solid outline shows the size of the $16\times 16$ receptive fields (RFs)
representation units in Layer 1.
The rectangle with the dashed outline shows the size of the $16 \times 26$ receptive
fields of the Layer~2 representation units.

The neurons $r_1^{(1)}$ -- $r_1^{(32)}$ in Fig.~\ref{fig_PE_level_1}
are the same neurons as those identified by $r_1^{(1,2)}$ -- $r_1^{(32,2)}$
in Fig.~\ref{fig_PE_level_global}.
The newly added Layer 3 receives input from all PEs patches from Layer 1.
Thus the receptive fields of representation units in Layer~3 are larger than
those in Layer~2.
In Fig.~\ref{fig_PE_level_global},
the middle component of Layer 1 corresponds to the module in Fig.~\ref{fig_PE_level_1}.

According to this model, endstopping phenomena occur 
in the prediction error neurons and reflects the reduced top-down
prediction error associated with the longer line segment.
Specifically, neurons
$e_{33}^{(2)}$ -- $e_{64}^{(2)}$
show reduced firing for the longer line on the right side of Fig.~\ref{fig_barStimuli_RaoBallard}
because of the two lateral patch modules in the first layer.
%
This is confirmed in the model by experiments which remove the feedback.
This possible explanation of end-stopping was proposed in \cite{Mumford1992}.

Equations~\ref{eq_RB_set}, \ref{eq_RB_predUpdateWithT}, and~\ref{eq_RB_predLearningWithT} can be extended to a multi-layer context
as shown in Section~\ref{sec_Purdue_predictiveUpdate}.

\textbf{This paragraph out of place.} The network was trained on several thousand natural image patches and the learning algorithm 
\textit{maximized
the posterior probability of generating the input data}.
Thus, this is a generative model.

\subsubsection{Problems with the Rao/Ballard model and future research}
From a neuroscience viewpoint, the model has one problem.
\begin{enumerate}
\item
This problem concerns the information transfer requirements.
It was stated earlier that a justification for predictive coding
was the decreased bandwidth needed to send equivalent information.
This is achieved in the feedforward direction.
However, the full representation is sent in the feedback direction retaining
the original bandwidth requirements.
This may be an issue that all hierarchical predictive coding models must
address.
\end{enumerate}

One idea for future research is to directly measure
the information transfer requirements (to address point 1)
on a spiking version of this model.
One could replace the continuous activation neurons with Poisson spiking neurons
(or perhaps a more efficient spike coding scheme).
By recording spiking events and measuring spike probabilities, we
could directly measure the information transferred in model.
We could compare the information transfer requirements for different coding schemes.

\subsection{Predictive coding and divisive input modulation}
\label{sec_SpratlingModels}
Spratling
\cite{Spratling2008a,Spratling2009a} used non-negative matrix factorization (NMF) to reformulate the \cite{rao1999predictive} model of predictive coding
to make it compatible with biased competition models of the brain.
A related technique, known as
divisive input modulation, offers promise of creating useful representations
of subparts and is related to non-negative matrix factorization.
\cite{Seung1999a} showed that NMF
could create a useful object representation of, say faces,
that consisted of local subparts resembling eyes, nose, and mouth.

More formally,
consider a set of images represented as an $n \times m$ matrix, named $\mathit{Images}$,
whose entries are all non-negative values.
The matrix
$\mathit{Images}$ contains $m$ images, $\boldsymbol{I}_1 \ldots \boldsymbol{I}_m$.
Each $\boldsymbol{I}_m$
holds the 
set of $N$ non-negative pixel values
for one image.
The goal is to learn representations for the images by refactoring 
the $\mathit{Images}$ matrix according
to the constraint that all entries in the factors are non-negative.
The refactoring takes the form below.

\begin{equation}
\mathit{Images} \approx W^T R
\end{equation}

\noindent
$W^T$ has dimension $n \times r$ and $R$ has dimension $r \times m$.
Think of $W^T$ as holding the feedback weights in a neural network and of $R$
as holding the transformed representation of each image within its columns.
Its non-negativity constraints cause the matrix to be refactored into a parts-based
representation.
This is because the factorization only allows additive combinations \cite{Seung1999a}.
Each column of $R$ is an internal representation of the corresponding image in $\textit{Images}$.
$r$ is the number of rows of $R$ and this determines the size of the representation.
For instance, if using the example given in Fig.~\ref{fig_PE_level_1},
$n$ would be 256 and $r$ would be 32.

\cite{Spratling2009a} has shown that by changing the operators used in the 
predictive coding formulas, 
a method related to NMF 
known as divisive input modulation
can be used 
which leads to some possible advantages over the \cite{rao1999predictive} approach.
Consider the revised
predictive update equations shown below \cite{Spratling2017b}.

\begin{subequations}
\begin{align}
\boldsymbol{\hat{I}} &= W^T \mathbf{r}\label{eq_Spratling_I}\\
\mathbf{e} &= \boldsymbol{I} \oslash \max\left(\epsilon_2, \boldsymbol{\hat{I}}\right)\label{eq_Spratling_predErr}\\
\mathbf{r} &\leftarrow \max\left(\epsilon_1, \mathbf{r}\right) \odot W\mathbf{e}\label{eq_Spratling_predictive_update}
\end{align}
\end{subequations}

\noindent
Eqn.~\ref{eq_Spratling_I} matches Eqn.~\ref{eq_RB_I}, except that it obeys the NMF constraint.
Eqns.~\ref{eq_Spratling_predErr} and~\ref{eq_Spratling_predictive_update} match Eqns.~\ref{eq_RB_predErr} 
and~\ref{eq_RB_predictive_update}, except that they use different operators.
Specifically, predictive error in
Eqn.~\ref{eq_Spratling_predErr} is calculated using divisive input modulation,
implemented by element-wise division ($\oslash$), instead
of subtractive inhibition.
For divisive input modulation, perfect reconstruction yields a value of one instead of zero.
For Eqn.~\ref{eq_Spratling_predictive_update}, elementwise multiplication ($\odot$) is used instead of addition
and this is an example of a multiplicative update rule \cite{Seung2001a}.
Expressions of the form ``$\max\left(\epsilon, \mathrm{Matrix}\right)$'' mean replace any values in $\mathrm{Matrix}$
that are smaller than $\epsilon$ with $\epsilon$.
For Eqn.~\ref{eq_Spratling_predErr}, this prevents division by zero
and for Eqn.~\ref{eq_Spratling_predictive_update} this prevents any of the $\boldsymbol{r}$ neurons
from becoming permanently non-responsive.
$\epsilon_1$ and $\epsilon_2$ are small positive numbers and example values to use
in simulations can be found in \cite{Spratling2017b}.

The weight update rule uses multiplicative update and is shown below \cite{Spratling2009a}.

\begin{equation}
W \leftarrow W \odot \left( 1 + \beta \boldsymbol{r} \left( \boldsymbol{e}^T -1 \right) \right)
\end{equation}

\noindent
Recall that when there is no prediction error, $\boldsymbol{e}^T - 1$ is zero and the right side
of the operator $\odot$ simplifies to one.
$\beta$ is a rate parameter.

\cite{Seung2001a} provide batch-based algorithms for NMF, one of which minimizes
the Kulback-Leibler (KL) divergence between the training and reconstructed images ($W^T \boldsymbol{r}$).
Whereas, the equations of Rao/Ballard are obtained by minimizing sum of squared error,
the equations above minimize KL divergence.




\cite{Spratling2017b} argues that his model has a number of advantages over
the original \cite{rao1999predictive} model.
First, it can be implemented using only  non-negative neural firing rates which makes it much easier to
create bio-plausible implementations.
Second, it is broadly compatible biased competition models \cite{Desimone1995} which also
makes it easier to create bio-plausible implementations.
Third, it has been argued that the proposed feedback connections of the Spratling model
are more biologically plausible than those in the \cite{rao1999predictive} model.

\subsection{The Friston Free Energy Model}
\label{section_Friston_model}
Following \cite{rao1999predictive}, \cite{Friston2003,Friston2005} developed a 
unified theory of sensory-based cortical function based on predictive coding
which estimated the mean and variance of predicted states.
This more general framework was
based on the ideas of hierarchical expectation maximization (EM)
and empirical
Bayes (a method to estimate priors from data).
This model is also biologically motivated along lines similar to \cite{rao1999predictive}, 
but is intended to address a broader range of
empirical evidence \cite{Friston2018a}.
The Friston model differs from \cite{rao1999predictive}, but still obtains 
the Rao-Ballard protocol.
As the Friston papers are mathematicaly difficult, 
there is an authoritative tutorial of
\cite{Bogacz2017} that explains the Friston model and is essential reading
to get started.
Although Friston formulates this model using a statistical framework,
our goal is to implement models of this type within deep learning frameworks that
scale to larger problems and, in practice, require fewer statistical assumptions.
We use the \cite{Bogacz2017} variant of the Friston model.

As mentioned earlier, 
building representations which capture causes of sensory input, such as the
shapes, reflective characteristics, arrangements and movements of objects, \cite[p.\ 151]{Spratling2017b}
is a type of
inversion process.
In the real world, 
this calculation may be 
intractable or impossible (ill-posed) because the inverse is not unique.
This can happen because of
nonlinear interactions among
the causes, for example visual occlusion \cite[p.\ 816]{Friston2005}.
This problem can be handled by estimating what the prior distribution should look
like. \textbf{Why? See Rao/Ballard, 1997}
\cite{Friston2005} uses empirical Bayes to obtain these priors.

Let us calculate the most likely value of a continuous quantity, given 
sensory evidence.
Following \cite{Bogacz2017},
we use a concrete
perceptual inference problem, namely, inferring the amount of a
food item in the environment from the intensity of 
its projection
on the retina.
The scalar variables $u$ and $v$ denote the pixel intensity and food amount, respectively.
The problem is formulated probabilistically, as $p(v|u)$.
This denotes the probability that the random variable $v$ (food quantity) has some value, given 
the pixel-intensity variable $u$ has some value.
We wish to compute the most likely value of $v$, given a specific value for the sensory input, $u$,
and the prior $p(v)$.
This is known as an MAP estimate.
Bayes's rule,

\begin{equation}
p(v|u) = \frac{p(u|v) p(v)}{p(u)} ,
\end{equation}

\noindent
lets us
find the most likely value of $v$, denoted $\phi$, by maximizing the log of the numerator 
(also known as \emph{negative free energy}\footnote{\cite[p.\ 821]{Friston2005} observes that ``minimizing the free energy 
is equivalent to minimizing a precision weighted sum of squared prediction error.''})
on the right-hand side (RHS) of the equation below, that is

\begin{equation}
\label{eq_neg_free_energy}
F = \ln p(u|v) + \ln p(v) .
\end{equation}

\noindent
Let us assume that $p(u|v)$ and $p(v)$ are normally distributed, where
$p(v) \sim\mathcal{N}(v_p,\sigma^2_p)$.
$v_p$ denotes the value of $v$ with the most likely prior probability and $\sigma_p^2$ denotes
the variance of the prior distribution.
Similarly, we assume $p(u|v) \sim\mathcal{N}(g(v), \sigma^2_u)$.
Equation~\ref{eq_neg_free_energy} can be expanded by substituting the formulas for these probability distributions.

Function $g(\cdot)$ predicts the raw sensory input (e.g., image pixel values)
from the representation by using a causal observation model.
This takes the form of
a deteministic,
usually nonlinear function of $v$ denoted $g(\cdot)$ and which may contain trainable parameters, $\theta$.

\begin{equation}
u = g(v; \theta)
\end{equation}

\noindent
$v$ is some representation of causes (e.g., amount of food),
$u$ is some representation of
the raw sensory input, and $\theta$ is a set of 
parameters (likely learnable) used to configure $g(\cdot)$.
To develop intuitions, 
one can think of $g(\cdot)$ as a deterministic causal process that
takes the state of the environment $v$ and maps it
to $u$.
An alternative interpretation is that $g(\cdot)$ is a function that
the brain can learn to approximate and the input to $g(\cdot)$ is
a representation, $v$, which resides in the brain but models causal
effects in the environment.
The output, $u$, of $g(\cdot)$ can then be viewed as a prediction of
what the sensory input should look like.
Although, in general $g(\cdot)$ is nonlinear, for this exposition we
assume that $g(\cdot)$ is linear.
For simplicity, we assume that $g(v;\theta)= \theta v$.

Our next steps for developing the scalar version of the free energy model are the following.
\begin{enumerate}
\item Develop equations to find the maximum of $F$ by performing gradient descent.
      Performing gradient descent on the resulting equations is called
      \emph{predictive activation update}.
      In appropriate contexts in can be called \textit{inference} (in contrast to learning).
\item Represent the equations in Step 1 as a neural network such that the trainable parameters map to
      weights or synapses in the network and the other parameters of the network map to neuron like elements (nodes) that
      can have activation levels.
\item Use $F$ to develop learning rules for the weights in the network.
\end{enumerate}

One way to maximize $F$ is to perform gradient ascent on $F$ with respect to the estimated
most likely value $\phi$.
The derivative of $F$ is
\begin{equation}
\label{eqn_negFreeEnergDeriv}
\frac{\partial F}{\partial\phi} = \frac{v_p-\phi}{\sigma_p^2} + \frac{\theta\phi-u}{\sigma_u^2} \theta.
\end{equation}

\noindent
Next, refactor this formula by defining two types of prediction errors involving $v_p$ and $\theta \phi$.

\begin{subequations}
\label{eqn_predictionErrors}
\begin{align}
e_p&\equiv\frac{\phi-v_p}{\sigma_p^2}\label{sub_predictionErrors_ep}\\
e_u&\equiv\frac{u-\theta\phi}{\sigma_u^2}\label{sub_predictionErrors_eu}
\end{align}
\end{subequations}

\noindent
With these prediction error definitions, Eqn~\ref{eqn_negFreeEnergDeriv} is rewritten as,
\begin{equation}
\label{eqn_negFreeEnerg_usingError}
\frac{\partial F}{\partial\phi} = e_u \theta - e_p ,
\end{equation}

\noindent
letting us express the problem of maximizing $F$ in terms of prediction error.
For a given parameter set, gradient descent on Eqn.~\ref{eqn_negFreeEnerg_usingError} can be used to search for an
optimal value of $F$.

\begin{figure}
\includegraphics[viewport=0.0in 7.0in 9.0in 10.0in,clip=true,scale=.65]{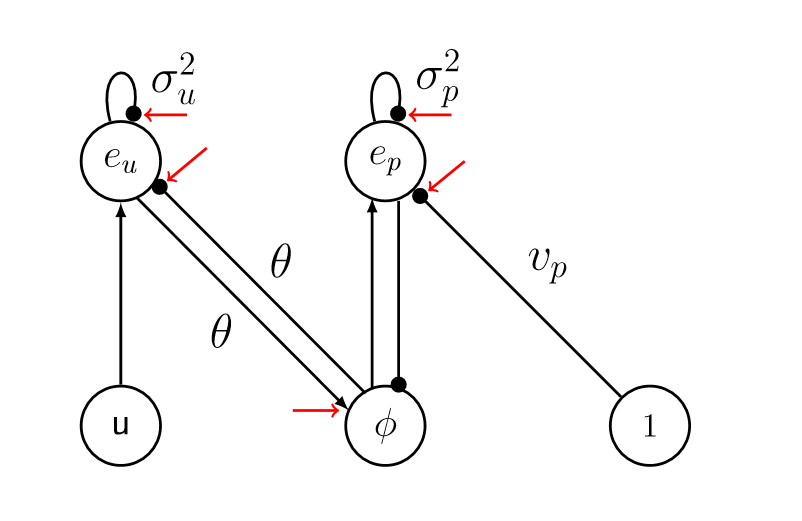}
\caption{A simple network that computes the most likely value of $\phi$ by calculating predictive error.
Red arrows indicate synapses where learning occurs.
Arrow heads indicate excitatory connections and circular heads indicate inhibitory connections.
Revised from \cite{Bogacz2017}.}
\label{fig_Bogacz_network_simple}
\end{figure}

Note that the neural model has been factored so that prediction errors are explicit nodes in the network
as seen in Fig.~\ref{fig_Bogacz_network_simple}.
The derivatives needed for gradient descent to update the nodes in the network are given as follows.
\begin{subequations}
\label{eqn_set_update_phi_scalar}
\begin{align}
\dot\phi &= \frac{\partial F}{\partial\phi} = e_u\theta - e_p\label{eqn_epsilon_phi_deriv}\\
\dot{e}_p &= \phi -v_p -\sigma^2_p e_p\label{eqn_epsilon_p_deriv}\\
\dot{e}_u &= u -\theta\phi -\sigma^2_u e_u . \label{eqn_epsilon_u_deriv}
\end{align}
\end{subequations}

\noindent
Eqn.~\ref{eqn_epsilon_phi_deriv} says that update to the representation
activations depends on the difference in prediction error for two consecutive layers.
Eqns.~\ref{eqn_epsilon_p_deriv} and \ref{eqn_epsilon_u_deriv} 
are obtained 
from 13a and 13b
by moving $e$ to the RHS, setting the derivative on the LHS to zero, 
and then simplifying
the RHS\@.
After convergence of gradient descent, the network reaches a state which maximizes $F$ with respect to the current 
parameter values.

Fig.~\ref{fig_Bogacz_network_simple} shows a possible neural implementation of this model.
The circled variables indicate continuous-valued, single neuron  activations.
The black lines ending with arrowheads and solid circles indictate excitatory and inhibitory neural  connections,
respectively.
The red arrows point to synaptic contacts where learning occurs.
The variables \textsf{u}, $\phi$, $e_u$, and $e_p$ are modeled as neural activations.
The labeled connections between the neurons $\theta$, $v_p$, $\sigma^2_u$, and $\sigma^2_p$ are
the values of the weights at the synaptic contacts and can be trained.
The trainable parameters have been made explicit by associating them with synaptic contacts.

The last step is to develop learning rules to train the network.
Learning equations for the five synapses are given below.
They are obtained by taking the derivative of $F$ with respect to the relevant synaptic parameter.

\begin{subequations}
\label{eqn_set_updateWts_scalar}
\begin{align}
\dot{v}_p        &= \frac{\partial F}{\partial v_p}        = e_p\label{eqVpriorLearnPE}\\
\dot{\sigma}^2_p &= \frac{\partial F}{\partial \sigma^2_p} = \frac{1}{2}\left(e^2_p - \frac{1}{\sigma^2_p}\right)\label{eqSigmaPLearn}\\
\dot{\sigma}^2_u &= \frac{\partial F}{\partial \sigma^2_u} = \frac{1}{2}\left(e^2_u - \frac{1}{\sigma^2_u}\right)\label{eqSigmaULearn}\\
\dot\theta       &= \frac{\partial F}{\partial \theta}     = e_u v\label{eqThetaLearnPE}
\end{align}
\end{subequations}

\noindent
Notice that all of the resulting learning rules are in some way functions of predictive error.
The prior probability, $v_p$, is not changed unless there is predictive error
related to prior (Eqn.~\ref{eqVpriorLearnPE}).
For Eqns.~\ref{eqVpriorLearnPE} and \ref{eqThetaLearnPE}, 
there is no learning when the predictive error is zero and learning increases with higher predictive error.
For Eqns.~\ref{eqSigmaPLearn} and~\ref{eqSigmaULearn}, the variance decreases when the predictive error is zero.

Fig.~\ref{fig_Bogacz_network_simple}
has two synapses with the same label, $\theta$.
Their separate weights are matched, in the sense that they each use the same learning rule, 
but not shared (they are different synapses).
The model can be simulated by running all of the dynamic equations simultaneously.

Running the network, given an input stimulus to be processed,
requires three steps as shown in the free energy algorithm below.


Performs network update (inference) and learning.

\noindent
\begin{enumerate}
\item Initialize the network with sensory input.
\item Perform \emph{predictive activation update} (inference) by gradient ascent using Equation Set~\ref{eqn_set_update_phi_scalar}
for some fixed number of integration steps (e.g., Forward-Euler), or until the node activations converge.
\item Apply learning rules in Equation Set~\ref{eqn_set_updateWts_scalar} to update the network weights.
\end{enumerate}

The algorithms of \cite{Wen2018a} and \cite{Han2018a}, described in Section~\ref{sec_thePurdueGroup}, show possible ways to do 
the above
in a deep learning framework for convolutional networks.

This model can be generalized to multiple variables and multiple layers.
The node dynamics and learning are governed by multidimensional versions of the rules
already described.
The generalized model uses multivariate instead of univariate Gaussian distributions.
Also, the distinction between $e$ and $e$ is generalized
to $e^l$ and $e^{l+1}$, where $l$ and $l+1$ stand for 
consecutive layers.
This is seen in Fig.~\ref{fig_predictiveElement} (b).

\begin{figure} \centering

\subfloat[]
{
\includegraphics[scale=.7]{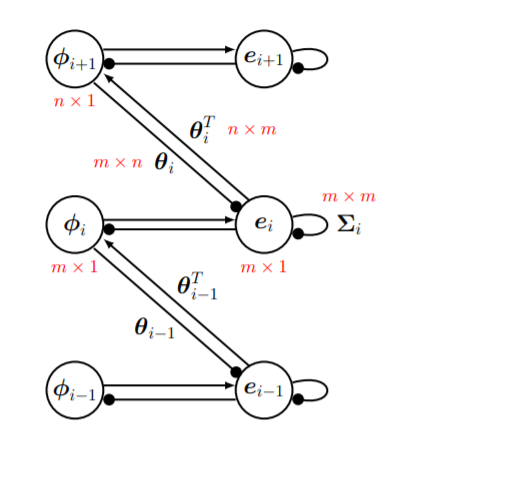}
} \qquad
\subfloat[]
{
\includegraphics[scale=.65]{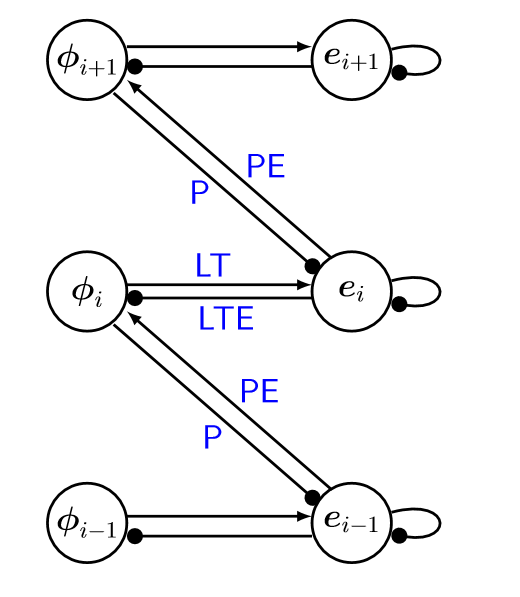}
}
\caption{(a) An RB protocol figure that captures some of the structure of Equation Set \ref{eqn_bogacz}.
(b) Same figure with RBP links made explicit.
Arrow-tipped links are additive and circle-tipped links are subtractive.}
\label{fig_RB_protocal_Bogacz}
\end{figure}

Predictive update dynamics are now specified using indexed equations across layers $l$, 
namely, a pair of coupled equations for each level in the hierarchy conforming to
Fig.~\ref{fig_predictiveElement}(b).
The $\boldsymbol{r}$ symbols in \cite{rao1999predictive} correspond approximately
to the $\boldsymbol{\phi}$ symbols in the equation below.
By extracting the couplings, we recover the Rao-Ballard protocol (see Fig.~\ref{fig_RB_protocal_Bogacz}).
\begin{subequations}
\label{eqn_bogacz}
\begin{align}
\dot{\boldsymbol\phi}^l     &= -\boldsymbol{e}^l + h^\prime(\boldsymbol\phi^l) \odot \left({\boldsymbol\Theta^{l-1}}\right)^T \boldsymbol{e}^{l-1}\label{eqn_bogacz_rep}\\
\dot{\boldsymbol{e}}^l &= {\boldsymbol\phi}^l - \boldsymbol\Theta^{\boldsymbol{l}}h(\boldsymbol{\phi}^{l+1}) - \boldsymbol\Sigma^{l}\boldsymbol{e}^l\label{eqn_bogacz_error}
\end{align}
\end{subequations}

\noindent
Eqn.~\ref{eqn_bogacz_rep} calculates the representation component in Fig.~\ref{fig_predictiveElement}(b)
and Eqn.~\ref{eqn_bogacz_error} calculates the predictive error for that layer.
Eqn.~\ref{eqn_bogacz_rep} has two additive terms on its RHS\@.
These approximately correspond to the information flowing on the incoming links to $\boldsymbol{r}^l$ in Fig.~\ref{fig_predictiveElement}(b).
Because of the couplings, we see that $\boldsymbol\phi^l$ receives input from both 
$\boldsymbol{e}^l$ and $\boldsymbol{e}^{l-1}$.
These correspond to the links {\small\textsf{LTE}} and {\small\textsf{PE}} in the Rao-Ballard protocol.
Similarly,  $\boldsymbol{e}^l$ receives inputs from $\boldsymbol\phi^l$ and $\boldsymbol\phi^{l+1}$ in the protocol.
These correspond to the links {$\small\textsf{P}$} and {$\small\textsf{LT}$} in the protocol.
Fig.~\ref{fig_RB_protocal_Bogacz} helps visualize this information.
Priors are represented on the {$\small\textsf{P}$} links.

Equation Set~\ref{eqn_bogacz} is numerically simulated by updating each equation across all layers
for each time step.
For reasons discussed in Section~\ref{sec_thePurdueGroup}, we call this a global update procedure.
Besides the links used in the Rao-Ballard protocol, we should also note the recurrent connections.
Eqn.~\ref{eqn_bogacz_rep} has a self reference in the second additive term on the RHS\@.
Eqn.~\ref{eqn_bogacz_error} also uses a recurrent connection in the third additive term on the RHS\@.

The next section considers the question of using these equations and the RB protocol as a guide to 
build a deep predictive coding network.
In a deep learning framework, the protocol constraints can be realized by ensuring that the both the 
$\boldsymbol{r}$ and $\boldsymbol{e}$ modules
are implemented as recurrent networks.
Also,
Equation Set~\ref{eqn_bogacz} is intended to operate on feature vectors and a deep learning model
will be applied to next-frame video prediction.
Therefore, must adjust the model so that it operates on multi-channel images.

In the spirit of \cite{rao1999predictive},
\cite{Friston2003,Friston2005} formulates this setup hierarchically
intending to capture different levels of representation that might occur
in the brain.
A given level provides priors to the immediately lower level
to support Bayesian inferences.
The model outline is shown below.

\begin{subequations}
\label{eqn_Friston_hierarchy}
\begin{align}
u &= g_1(v_1, \theta_1) + \delta_1 \label{sub_Friston_hierachy_a}\\
v_1 &= g_2(v_2, \theta_2) +  \delta_2  \label{sub_Friston_hierachy_b}\\
v_2 & = \cdots
\end{align}
\end{subequations}

\noindent
The higher-level representations, $v_i$, predict 
the state of immediately lower-level representations, $v_{i-1}$.
$u$ at the lowest level denotes observed values at the sensory receptor.
Each $g_i$ function is deterministic with its own parameter set $\theta_i$.
The $\delta_i$'s add noise to the predictions.

By adding assumptions about the noise sources being Gaussian with
covariance $\Sigma_i$,
we can reformulate Eqns.\ \ref{eqn_Friston_hierarchy} probabilistically.
The equation below expresses the likelihood of $v_i$ given the state
of $v_{i+1}$.

\begin{equation}
p(v_i | v_{i+1}; \theta) = N(v_i : g_{i+1}(v_{i+1}, \theta_{i+1}), \Sigma_{i+1})
\end{equation}

\noindent
$v_i$ and $v_{i+1}$ are now vectors of continuous variables and
$v_i$ is sampled from a multi-variate Gaussian distribution with
mean $g_{i+1}(v_{i+1}, \theta_{i+1})$ and variance $\Sigma_{i+1}$.
The representation at level $i+1$ provides priors for level $i$
providing top-down expectations to the lower layer.

Before proceeding, let us review EM in a nonhierarchical context.
The EM learning algorithm is iterative 
with two steps in the iteration loop: the E-step and the M-step.
The E-step uses the recognition equation below to assign probability values to the causes, $v$,
of a given input, $u$, using the current parameter values, $\theta$.
\begin{equation}
\label{eqn_EM_recognition}
p(v|u;\theta) = \frac{p(u|v;\theta) p(v;\theta)}{p(u;\theta)}
\end{equation}

\noindent
Using Bayes's rule, this equation computes the probability of
a cause given the current input and the current parameterization
of the model, $\theta$, which will be adjusted in the M-step.
The rule inverts the generative formula $p(u|v;\theta)$,
which predicts the input distribution.
If the value of formula \ref{eqn_EM_recognition} can be computed tractably, 
the model is said to be invertible.
Otherwise, it is noninvertible \cite{Dayan01}
and relates to the ill-posed nature of sensory input mentioned at the start
of this section.
In the context where the model is not invertible,
we introduce an approximate recognition distribution written

\begin{equation}
q(v ; u, \phi)
\end{equation}

\noindent
which uses a new parameter set, $\phi$,
called recognition parameters.

We of course need a way to estimate this approximate distribution.
Predictive coding is a form of EM that finesses the challenge of
finding the parameters for complicated recognition densities.
In a predictive coding framework, the $\phi$ parameters
``are dynamically encoded by the activity of units in the brain'' 
\cite[p.\ 1339]{Friston2003}.

The M-step uses the results of the E-step to update the model parameters, 
and this is how learning occurs.
In traditional EM,
the specific update equations depend on the details of the 
particular causal model used.
A common didactic example to illustrate EM is to use a Gaussian mixture model 
to generate data which is to be clustered.
In this case, the parameter set consists of: 1) the mixture proportions;
2) the mean vectors of each distribution in the mixture; and, 3) the parameters
for the variance of each mixture distribution.
The specific details of such causal models can be arbitrarily complicated.
Predictive coding is a version of EM that avoids the challenges
of formulating and parameterizing specific causal models.

According to \cite{Friston2005}, EM is ``a useful procedure for density estimation
that has direct connections to statistical mechanics'' [p.\ 820].

\begin{figure}
\includegraphics[scale=.55]{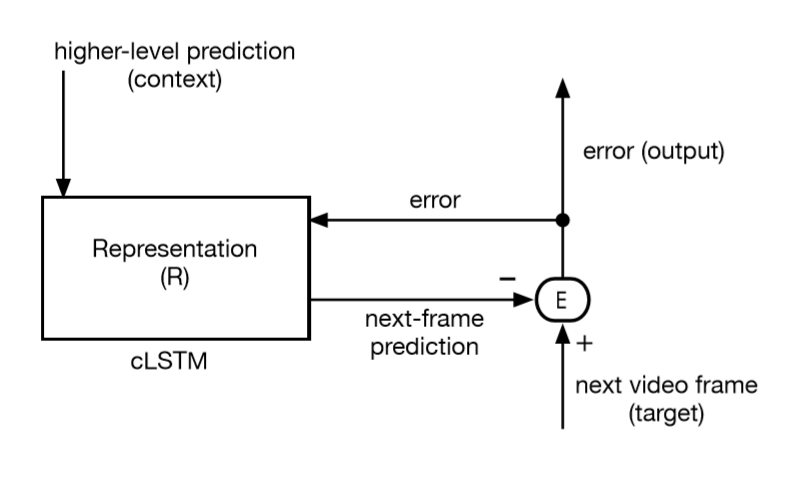}
\caption{Information flow in the lowest PredNet layer
(training mode) where the input is a ground truth video
frame.
The representation, $\mathsf{R}$, and prediction error components, $\mathsf{E}$, are recurrently connected.}
\label{fig_prednet_layer}
\end{figure}

\section{Predictive coding in a deep learning framework}
\label{sub_LotterModel}
\cite{Chalasani2013a} may have been
the first to use the term ``deep predictive coding networks.''
However, the model of \cite{Lotter2017}, named PredNet,
may be the earliest predictive coding model implemented using a deep
learning (DL) framework.
Implementing a predictive coding model using a DL framework has a number
of potential advantages over implementing the direct mathematical formulations
described earlier.
DL frameworks are very mature, versatile, and efficient.
Consequently,
they should make it easier to build and study predictive coding models,
with the only complication being their ability to handle cross-layer feedback connections.
Second, models using DL frameworks scale to very large architectures
with over one hundred thousand parameters.
This has not been achieved using traditional implementations of predictive coding.
Third, deep learning architectures allow the use of large learning
modules (such as LSTMs) that can cope with more relaxed statistical
assumptions and, thereby, operate in more general situations. 
For instance, they allow us to drop strong assumptions such 
as Gaussian priors.
Since PredNet is the first predictive coding model implemented
using a DL framework, we shall discuss this model in detail.

PredNet falls into the second category of models mentioned in the introduction.
That is, it predicts prediction error.
It is applied to the task of next-frame video prediction.
Because PredNet manipulates video data,
the representation modules in this model consist of 
convolutional LSTMs (cLSTMs)\@ \cite{Shi2015a}\footnote{A convolutional LSTM is an 
LSTM whose internal data structure is a multichannel image.}.
LSTMs are recurrent networks with trainable gating operations to help them remember or forget
different types of relevant or irrelevant past information when performing sequence processing.
The internal data structure that they manipulate is a feature vector and this is suitable for
applications like natural language processing.
cLSTMs are a modification of LSTMs that use multi-channel images as their internal
data structure in place of feature vectors.
Accordingly, cLSTMs
replace gate operations based on affine weight multiplication (used in regular LSTMs)
with convolutional gate operations that apply to multi-channel images.
This allows them to develop useful representations of image sequences such as video.

Fig.~\ref{fig_prednet_layer} shows a PredNet predictive element for the lowest layer of the model
where the representation module on the left is implemented as a cLSTM\@.
By using the cLSTM, PredNet's architecture generalizes predictive coding models
to perform more complex tasks than, say, that of \cite{rao1999predictive}.
Note that the output of the representation module projects to the error calculation module,
which sends its output back to the representation module.
The model learns to predict the next frame in a video (target) by comparing its prediction with
the target frame and using the prediction error as a cost function.
Since Fig.~\ref{fig_prednet_layer} does not show how the feedforward and feedback connections link to the
next higher layer, we cannot determine whether it is a model that predicts prediction error.
At this point, it is generic as a predictive coding model.

Although, PredNet is inspired by
predictive coding principles, its design diverges fundamentally
from the models of \cite{rao1999predictive}, \cite{Friston2005}, and \cite{Spratling2009a}.
The most noticeable difference is that it uses cLSTMs to implement the representation modules instead of
layers of neurons.

\begin{figure} \centering
\subfloat[]
   {
   \includegraphics[scale=0.5]
   {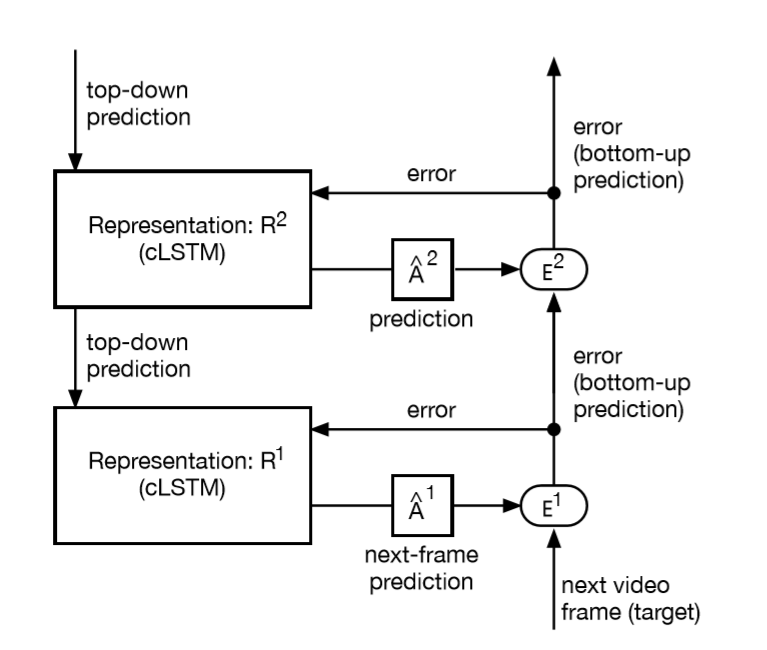}
   } \qquad
\subfloat[]
   {
   \includegraphics[,scale=0.5]
   {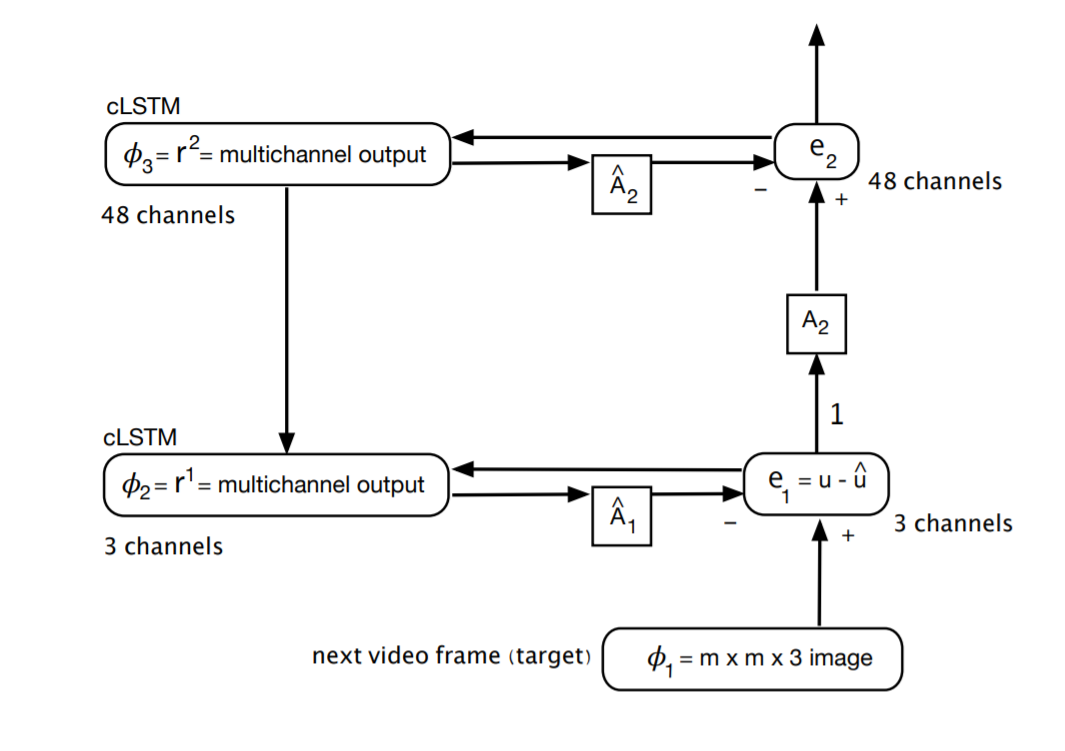}
   }
\caption{\small Two views of the Lotter et al.\ model.
(a) Simplified figure from \cite{Lotter2017}.
(b) Refactored figure to compare \cite{Lotter2017} with the earlier predictive coding models.}
\label{fig_twoPossibleModels}
\end{figure}

A more subtle and fundamental difference between PredNet and the earlier models
is that the intermodule connectivity in PredNet diverges from the previous
models we have examined. 
Specifically, PredNet does not use the RB protocol.
This cannot be readily seen in Fig.\ref{fig_prednet_layer}, but is more apparent in
Fig.~\ref{fig_twoPossibleModels}(a) which shows a two-layer version of the model.
The module interconnectivity pattern differs from the RB protocol.
For example, the second-layer representation projects to the first-layer representation,
whereas, if using the RB protocol, it would project to the first-layer \textit{error}.
Similarly, if using the RB protocol,
the first-layer should project to the second-layer \textit{representation}.
Instead it projects to the second-layer error.

Let us examine intuitively how PredNet works when a new video frame is given to the network.
We start with the lowest layer.
The representation layer ($R^1$) generates a prediction for the next video frame on the basis of the knowledge
encoded in its weights.
The prediction is sent to the error module ($E^1$) via the convolution module ($\hat{A}^1$) which
modifies the dimensions coming from $R^1$ so that it matches the error module.
The error module also receives the target, which is the actual video frame that appears next in the
sequence.
The error module computes prediction errors by performing two subtractions.
The prediction is subtracted from the target (e), and the prediction error is sent back to
the representation, as well as to the next higher layer.
Similar to the previous models
\cite{rao1999predictive,Bogacz2017},
the error module is subtractive.

Learning adjusts the weights for the convolutional filters
as well as the weights for the internal gates within the cLSTM\@.

The error modules, $\boldsymbol{e}_i$, in the model are subtractive as in the previous models
\cite{rao1999predictive,Bogacz2017}.
When the representation and error modules are combined, they form a 
network layer as shown in Fig.~\ref{fig_prednet_layer}.
These layers can be stacked and 
a two-layer version of the model is shown in Fig.~\ref{fig_twoPossibleModels}(a).
In this figure, the inter-layer connectivity becomes apparent.
The network weights are trained by backpropagation where the cost function is a weighted
sum of the prediction errors generated by the $\boldsymbol{e}_i$ modules for 
all of the layers.
For the four-layer version of their network, they used two versions of
a training loss function, $L_0$ and $L_\mathrm{all}$, which are based on how 
the $\boldsymbol{e}_i$ are weighted.\footnote{In the \cite{Lotter2017} model,
these loss weightings are static throughout training, in the earlier models discussed
they are dynamically weighted according to the calculated precision.}
For the $L_0$ loss, the four $\boldsymbol{e}_i$ are weighted as $\left[1, 0, 0, 0\right]$.
The $L_\mathrm{all}$ loss uses $\left[1, .1, .1, .1\right]$ which gives a small weighting to each
of the higher layers.
The model performed slightly better when the $L_0$ loss was
used, which means that incorporating loss in the high-level modules, very surprisingly,
reduced performance quality.

Once trained,
their model predicts the next frame in a video sequence after seeing 
some number of
previous frames in the sequence (e.g., 10)\@.
The model uses a hierarchy of predictive coding layers with bottom-up (feedforward) and
top-down (feedback) connections.
The feedforward connections convey prediction errors to the 
error module in the
next higher layer and the feedback connections convey
representation updates to the
representation module in 
the previous layer
(see Fig.~\ref{fig_twoPossibleModels}(a))\@.
The prediction at the lowest layer is intended to predict and match the next frame in the video sequence.
Characteristic of deep learning, the model uses convolutional and recurrent (cLSTM) modules.
As such, it is an important contribution that merges deep learning with predictive coding.

First, 
as seen in Figure~\ref{fig_twoPossibleModels}a,
the feedforward connections project to prediction error neurons instead
of representation neurons.
Second,
the feedback connections project to representation neurons and not prediction error
neurons.
Both of these properties depart from
the Rao-Ballard protocol used in the above mentioned models.

Specifically, the prediction error for a lower layer is sent directly
to the error calculation for the next higher layer
in contrast to the Rao-Ballard protocol where prediction error would be sent to the next
higher representation layer.
The representation for a higher layer is sent to the representation for the previous layer
in contrast to the Rao-Ballard protocol where it would be sent to the error calculation
for the previous layer.
Thus, PredNet diverges from the earlier predictive coding models.

We now present the training cycle for the architecture.
This can be compared with
the Free Energy algorithm in Section~\ref{section_Friston_model}.

Performs network activation update during training.
Operates on one frame of input in a video stream.
Model attempts to predict the next video frame and calculates prediction error.
An activation update step is described below.
\begin{enumerate}
\item Input current video frame.
\item Perform top-down update sweep on the representations.
\item Perform bottom-up update sweep on the predictive errors.
\end{enumerate}

Weight update is controlled by backpropagation on the mean absolute error (MAE).
MAE is calculated by summing the prediction errors across layers and the previous history
ten steps into the past.
Backpropagation is applied to reduce MAE\@.

The \cite{rao1999predictive} model showed how larger spatial context from a higher layer can 
provide information to a lower layer to reduce prediction error.
Although the 
\cite{Lotter2017} model provides higher-level feedback from a 
higher layer representation
module to the previous layer representation module, the nature of the information provided is different.
The representation module in the second layer of the Lotter model (see Fig.~\ref{fig_twoPossibleModels}(a))
\emph{appears to be trained} (more explanation shortly)
to predict the prediction error generated by the first layer.
That is,
it is not
being trained to predict the input to the lower layer, 
as was the case for previous models.
This means that the nature of the higher-level information provided is different.
It provides information to predict the prediction error and it is not obvious how
this is useful in actually reducing prediction error.
There is a missing link in the theory.
Heuristically, it seems counter intuitive and \cite{Lotter2017} do not provide
a rationale for making this choice.
Although there is evidence that the model gives good performance,
there is not a well articulated theory of why this model functions.

Let us analyze Fig.~\ref{fig_twoPossibleModels}a to try to develop a theory of how the model functions.
We focus on how the two-layer model should be better than the one-layer model.
When using the $L_0$ training loss,
the error value $\boldsymbol{e}_2$ is not part of the training loss function, so
learning in the representation at Layer 2 only reduces the loss $\boldsymbol{e}_1$.
The hedge ``appears to be trained'' used in the previous paragraph relates to this.
The network architecture suggests that level-two representation is reducing higher-order
error, but the $L_0$ loss function contradicts this.
Since $\boldsymbol{e}_2$ is not influencing the training,
The information for the backpropagated weight updates,
originating with $\boldsymbol{e}_2$, flows in the opposite direction
that the arrows point.

\textbf{Hypothesis}: If we cut the link labeled ``1'', it should have a negligible effect on performance.
If this hypothesis turns out to be true, then the higher-level
predictive error calculations perform no significant function.
This would mean that
the Lotter model is not really
a predictive coding network, and that its principle of function 
is similar to that of a traditional deep network.
Specifically, it is a hierarchical cLSTM network that uses sum-of-squared error
loss at the lowest layer.

Although not explicitly shown in the figures,
pooling is used on the upward links
and upsampling is used on the downward links between consecutive layers.
This implements some form of hierarchical spatial context but, as mentioned above,
it is difficult to heuristically interpret by virtue of it treating prediction error 
as the higher-level representation.

The unit activations in the model do not have obvious probabilistic interpretations
but the model does appear to be generative in the way that it creates its 
next frame predictions.

\subsection{Modifying the Lotter Model to use the Rao-Ballard Protocol}
\begin{figure} \centering
\includegraphics[scale=0.6]{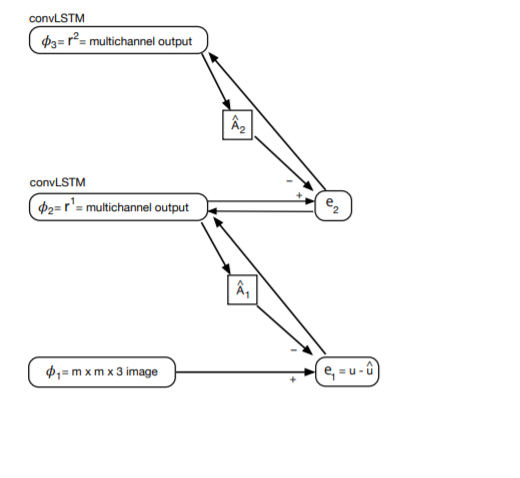}
\caption{\small A deep architecture which more closely matches the earlier predictive coding models
because it obeys the Rao-Ballard protocol.}
\label{fig_revisedLotterModelCompatibleWithRao}
\end{figure}

Because of the novelty of the \cite{Lotter2017} model's design,
it is useful to compare the original \cite{Lotter2017} model with
a version that obeys the more established Rao-Ballard protocol.
This comparison has not yet been performed.
Thus,
we propose a novel revision to \cite{Lotter2017} so that 
its module connectivity conforms to the Rao-Ballard protocol
which can be used as a baseline to evaluate the original model.
Our revised deep-learning-based architecture, which is heuristically
similar to the \cite{rao1999predictive} or \cite{Friston2005} design, is shown in 
Fig.~\ref{fig_revisedLotterModelCompatibleWithRao}.
It should be compared with Fig.~\ref{fig_twoPossibleModels}(b).
The original \cite{Lotter2017} model uses top-down feedback to directly update
the next lower representation.
The version that incorporates the Rao-Ballard protocol uses the top-down information
as a prediction that enables prediction error to be computed.
The loss function used for training should be $L_\mathrm{all}$.

Fig.~\ref{fig_revisedLotterModelCompatibleWithRao} depicts
a novel model not yet studied.
It differs form the original \cite{Lotter2017} model in two ways.
First, vertical links between the representation components and between the error components are
dropped.
Thus, there are no direct feedback connections between representations in consecutive layers.
Direct feedforward connections between error modules in consecutive layers is also dropped.
Second, horizontal links between the representation and error components are added.
So, the representation in a given layer receives input from the error module in the current layer
and also from the error module in the previous layer.
This figure should also be compared with Fig.~\ref{fig_predictiveElement}(b).
This model 
matches the original heuristic intentions of the Rao-Ballard protocol
although we do not supply a formal probabilistic interpretation.

\subsection{Potential for bioplausibility}
Recent simulation experiments have been conducted in an effort to
support the biological plausibility of the model \cite{Lotter2018a,Watanabe2018a}.

The Lotter et al.\
model predicts several neuroscience 
phenomena \cite{Lotter2018a,Watanabe2018a}
including the time-course of neural firing rates after stimulus onset, 
end-stopping and other phenomena related to non-classical receptive fields, and 
a number optical illusions.
Regarding firing rates,
neurons exhibit an on/off temporal dynamics
in regard to presenting and removing visual inputs
in some parts of the visual cortex (area V2). 
This applies to stimuli to which the neuron is tuned.
Such neurons have a low baseline firing rate before stimulus presentation.
After stimulus onset, the neuron's firing rate transiently increases and then returns
to baseline.
After returning to baseline,
the same neuron's firing rate will transiently increase when that same stimulus is turned off.
This naturally fits a predictive coding interpretation where the neuron becomes active in
response to prediction failures (It predicts constancy in the stimulus).
\cite{Lotter2018a} have shown that artificial neurons within prednet can exhibit this
behavior at a population level (but not at an individual neuron level).
They presented PredNet with static inputs of a range of naturalistic objects at time-step zero.
Then they removed the stimulus at time-step six.
When averaged over the stimuli and units for a given layer, the population response
matched the response dynamics of primate V2 neurons.
This is consistent with the interpretation that the population dynamics of PredNet were operating
within a predictive coding regime.
It also suggests analogies between the function of PredNet and the visual cortex.

\cite{Lotter2018a} and \cite{Watanabe2018a} have shown that PredNet can account for
some visual illusions.
These include illusory contours, the flash-lag illusion, and illusory motion in static images.
\cite{Watanabe2018a} showed that PredNet can predict illusory motion.
This phenomenon arises when motion is perceived  in static 
images.
For examples, see Akiyoshi's illusion pages \url{http:www.ritsumei.ac.jp/\~akitaoka/index-e.html}.
These experiments were conducted because previous papers suggested that
predictive coding offered an explanation of motion illusions 
\cite{Notredame2014,Nour2015,Shipp2016}.
One important aspect of this study is the methods they introduced to analyze
effects of the predictive coding mechanism within the network.
Their most impressive analysis was an optical flow analysis used
to guide image \cite{Kanada_1981a}
registration for predicted images as a means measure the predicted
motion in consecutive image frames.

\subsection{Additional potential for bioplausibility}
\label{sec_potential_bioplausibility}
As stated earlier, other representations can be used in place of the convolutional 
LSTM\@.
The convolutional LSTM can be replaced by a reduced-gate network which, under some circumstances,
gives comparable performance with fewer parameters
\cite{DBLP:journals/corr/abs-1810-07251}.
It can also be replaced by an Inception-based LSTM \cite{hosseini2019inception}.

More biologically plausible alternatives are possible.
\cite{Costa2017a} developed an LSTM that uses subtractive gating (because subtractive
inhibition exists in the brain, Fig.~\ref{fig_Bogacz_network_simple}),
instead of multiplicative gating.
If this subtractive LSTM were extended to be convolutional, then it would be a candidate
to be used in the \cite{Lotter2017} model to increase its bioplausibility.
\textbf{Possible research idea.}
In another vein, \cite{Spratling2010a}  and Section~\ref{sec_SpratlingModels} explored using divisive inhibition (versus
subtractive inhibition) as a more bioplausible means of
implementing predictive coding.
This could be incorporated in \cite{Lotter2017}.

\subsection{Significance of the Lotter model}
\label{sec_Lotter_significance}
The \cite{Lotter2017} model may be the first predictive coding model implemented
within a deep learning framework.
This
model is much larger and more comprehensive than using a formally specified statistical model
(e.g., \cite{Friston2005}).
It moves in the nonparametric direction, dropping
specific statistical assumptions like Gaussian priors.
The free-energy principle is as compatible with deep learning models as it is with formally 
specified statistical models.
\cite{rao1999predictive} used a neural network implementation but because they did not have
access to a deep learning development environment,
they were limited in the size, complexity, and variety of the models they could study.
\vspace{12pt}

\subsection{Revising the Lotter Model}
\label{sec_hybridModel}
\cite{Lotter2017} used an architecture which forwarded prediction errors 
as a prediction for the next level in the hierarchy.
The \cite{rao1999predictive} model did this also.
They differ in where the representation information for a given layer $l$
is sent.
For \cite{rao1999predictive}, the representation information is used to compute
prediction error for Layer $l+1$,
whereas \cite{Lotter2017} uses that information to compute the prediction error
for Layer $l$.

The \cite{Lotter2017} 
representation acquired in Layer $l+1$ is trained to reduce only prediction errors from
the previous layer.
It seems that the \cite{Lotter2017} model is trained to acquire a hierarchy of
high-order errors.\footnote{A loose analogy to the Taylor series might                  
provide intuition here. Think of the higher layers as corresponding to the higher-order
derivatives of the Taylor expansion. It is not clear that these higher order
derivatives would be useful in a representational context.} 
The \cite{rao1999predictive} design seems better suited to acquire a representation hierarchy.

\noindent
\textbf{Open questions and possible further experiments:} 
\begin{enumerate}
\item
Does the Lotter model give state-of-the-art video prediction performance?
There do not appear to be comparisons with other approaches.
\item
Quantifying the effect of the number of layers.
The \cite{Lotter2017} model was applied to two data sets using five (face rotations data set)
and four layers (KITTI traffic data set), respectively.
However,
their short paper did not quantify the effect of adding more layers.
Experiments should be done to confirm that adding more than one layer does indeed
improve performance and the potential improvement should be quantified.
Without these experiments, we do not know for a fact that the larger spatial context 
of higher layers improves performance 
on these data sets.
If this is not confirmed, the predictive coding theory associated with the model
is meaningless.

\item
Experiments with moving MNIST \cite{DBLP:journals/corr/SrivastavaMS15}, as well as our own experiments
with the Lotter model, show that it is easier to predict the movement of the digits
than it is to predict their shape (even though digit shape is fixed).
One could say that the digit shapes are low-pass filtered but
this is not very informative.
It only says there is information loss in the system.
Why is it so hard for this version of the architecture to acquire knowledge of digit shape?
What needs to be added?
Maybe position information is less memory intensive and that some form of auto-encoder functionality
should be added to encode shape.

\item
Analysis of prediction accuracy.
It was pointed out earlier that the \cite{Lotter2017} model lacks a clear theory
of why the model functions.
We must understand what predictive information is learned in the representation module.
In regard to movement (but not shape),
measuring optical flow (using the CV2 library) between the representations of consecutive images may help.
Our own experiments lack this analysis.
Four types of comparisons would help.
\begin{enumerate}
\item An optical flow comparison between predicted and target images.
      This would give detailed information about the mismatch between the predicted
      and target images and
      could serve as a cost function that is more informative than
      the SSE cost function.
\item An optical flow comparison between the previous and current
      target.  This would give ground truth information about how the image
      changes.
\item Comparison between the previous target and current prediction.
      Combining this with information from item b,
      gives information about how the predictions go wrong.
\item Visualizations of within-channel optical flow in the representation module could reveal quantitative
      information about the properties of the internal representations. 
      Is optical flow the same for all channels?
\end{enumerate}

\item
Increasing bio-plausibility.
Many papers offer hypotheses 
\cite{Bastos2012a,Shipp2016}
\textbf{citations} about how predictive coding can be
implemented in minicolumns.
We should compare these with PredNet and similar models.
If we built a subtractive convLSTM, would that work in the Lotter model?
Which is more compatible with the minicolumn hypothesis: the original Lotter model,
or the model modified to use the Rao-Ballard protocol?


\item
Improving Prednet by using trainable upsampling.
The top-down feedback in the Prednet implementation uses the Keras \texttt{UpSampling2D} API\@.
This just resizes the image.
Performance may be improved by instead using the \texttt{Conv2DTranspose} API
which has trainable weights.
This allows search to find
the best upsampling to reduce the cost.
This increases the parameter budget and may require more training data
to prevent overfitting.
\end{enumerate}



\section{Predictive update can improve classification}
\label{sec_improving_feature_hierarchies}
Except for \cite{Spratling2017b},
the previously discussed models have not been used for classification.
Predictive update (inference) is the second step in Free-Energy Algorithm~\ref{freeEngControl}.
A pair of studies has shown that predictive update can improve the performance, 
of convolutional feature hierarchies \cite{Wen2018a,Han2018a}, including
better state-of-the-art classification accuracy, faster convergence
time, and the need for fewer layers.

When using a feedforward hierarchy of feature representations,
the presumption
is that the representation in each layer within
the hierarchy should use different features.\footnote{With some exceptions such as ResNet
layers which are bypassed because of skip connections.}
Consider two consecutive layers, $l$ and $l-1$, which are
fully and bidirectionally connected, have the same number of units, and also have identical linear activations,
that is, the representations are identical.
If the feedback connections from $l$ to $l-1$ implement the identity transform,
then layer $l$ can perfectly reconstruct the representation in $l-1$.
In this case, there is no prediction error to trigger learning,
and the higher-layer representation is stable \cite{WhittingtonBogacz2017}\footnote{There are empirical results
\cite{Han2018a} described later in this section that the representation converges
under predictive update.}
when
identical to the lower-layer representation.
When identical, there is no need to have 
both layers in the hierarchy.
Additional constraint must be added to the system, most likely at layer $l$, to make the representations
in consecutive layers different, so that deep learning is useful.
That is,
some constraint must be imposed on layer $l$ so that,
while it uses layer $l-1$ to create a representation, it is forced to create
a different (higher-level) representation.

To set up a baseline predictive coding model within a deep learning framework for classification,
consider a feedforward convolutional architecture with $n$ layers.
Now extend 
the standard feedforward convolutional architecture to include predictive coding by adding feedback connections to each layer.
The feedback connections originating in layer $l$ use the representation in layer $l$
to reconstruct and predict the representation in layer $l-1$.
Prediction errors are communicated forward
from layer $l-1$ to layer $l$
and these errors guide learning (minimize reconstruction errors by gradient descent on error)
of the feedforward and feedback weights to reduce future
reconstruction errors.
We will call this the basic PredCode architecture.


\subsection{The Purdue group and predictive activation update}
\label{sec_thePurdueGroup}
A pair of studies \cite{Wen2018a,Han2018a} from Purdue
introduced the above basic PredCode architecture.
They
added
predictive activation update, similar to that described in Section~\ref{section_Friston_model},
into deep convolutional architectures.
They compare identical deep learning architectures with and without
predictive activation update so that the effects can be assessed.
It was found that predictive activation update yields faster learning convergence and better
classification accuracy as compared to the plain deep-learning versions of these networks
when tested on data sets like CIFAR, MNIST, and SVHN, regardless of whether the predictive activation update
is global or local.
Their results also show that predictive activation
reduces the need for extremely
deep feature hierarchies.
These predictive coding networks were less than ten layers deep.
The improved prediction accuracy remained even when the number of parameters between
the plain and PCN versions were held constant.

\subsubsection{Adding predictive update to a baseline convNet}
\label{sec_Purdue_predictiveUpdate}
The first study mentioned above, \cite{Wen2018a}, used a global version of predictive update.
This method propagates the updates throughout the hierarchy for each update cycle.
They studied visual classification problems on the data sets mentioned above and
started with a baseline VGG-like CNN classification architecture 
\cite{Simonyan2015a} 
that was upgraded to use global average pooling \cite{LinCY2013a}.
This was called the plain CNN\@.
The enhanced network was called predictive coding net (PCN)\@.
PCN was directly compared to the plain CNN and achieved the improved results described
at the beginning of Section~\ref{sec_thePurdueGroup}.

To add predictive update functionality to the plain CNN, 
\cite{Wen2018a} introduced recurrent processing for each 
pair of consecutive layers
of the network like that suggested in Section~\ref{sec_improving_feature_hierarchies}
to build the basic predictive coding architecture.
They derived their equations for a fully connected network although their actual implementation
used a convolutional network.
Working within the framework of a fully connected network with linear layers,
they started with the equations below to compute prediction error:

\begin{subequations}
\label{eqn_Purdue_error}
\begin{align}
\boldsymbol{\hat{r}}^{l-1}(t) &= W^{(l, l-1)} \boldsymbol{r}^l(t)\label{eqPredictionPurdue}\\
\boldsymbol{e}^{l-1}(t) &= \boldsymbol{r}^{l-1}(t) - \boldsymbol{\hat{r}}^{l-1}(t)\label{eqPredErrPurdue} .
\end{align}
\end{subequations}

\begin{figure}
\includegraphics[scale=.6]{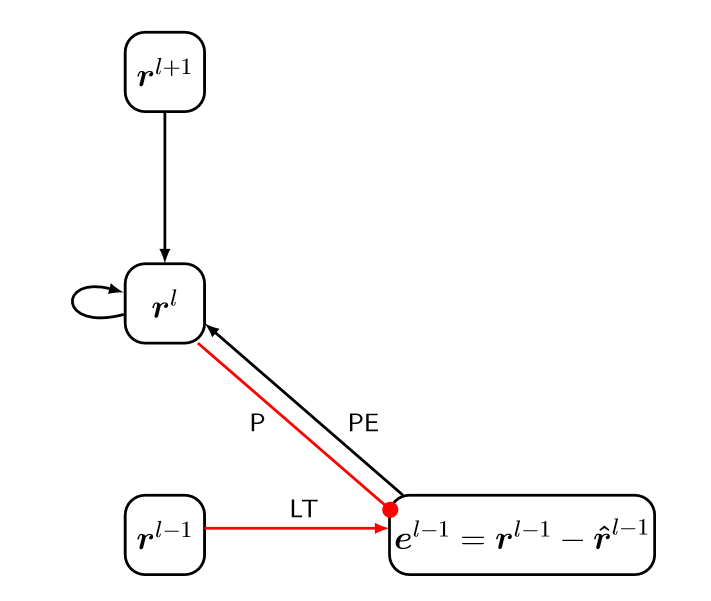}
\caption{Data flow for one cycle of predictive activation update for layer $l$ in the \cite{Wen2018a} 
and \cite{Han2018a} models.
It differs
from the Rao-Ballard protocol in the way that it uses top-down information. 
The red links highlight the predictive error calculation.}
\label{purdue_group_dataflow}
\end{figure}

\noindent
The above equations are consistent with the \cite{rao1999predictive} Eqns.~\ref{eq_RB_I}
and~\ref{eq_RB_predErr}, except that they are formulated to apply in a hierarchical context.
$W^{(l,l-1)}$ denotes the feedback weight matrix originating in layer $l$ and
projecting to layer $l-1$.
Unlike Eqn.~\ref{eq_RB_I}, it is not required that $W^{(l,l-1)}$
be equal to $(W^{(l-1,l)})^T$.
$\boldsymbol{e}^{(l-1)}$ is the prediction error for layer $l-1$, 
where $\boldsymbol{\hat{r}}^{(l-1)}$ is the feedback prediction from layer $l$.
By examining the communication within and between layers $l$ and $l-1$, we see that
it uses part of the Rao-Ballard protocol, namely, the P and LT links (Fig.~\ref{fig_predictiveElement})\@.
Later, we will see that it also uses the P link as seen in Fig.~\ref{purdue_group_dataflow}.
However, it does not use the LTE link.
When comparing with Equation Set~\ref{eq_RB_set},
we see that the update equation has not yet been specified.

To obtain the recurrent update equations, we take the derivative of the loss with respect
to the representation activations for a particular layer, $l$, to set up gradient descent.
Following Eqn.~\ref{eq_RB_ssquaresLoss},
the loss for layer $l$ is given as:

\begin{equation}
J^{l}(t) = || \boldsymbol{e}^{l}(t) ||^2
\end{equation}

\noindent
The derivative of the loss with respect to changes in the activation level of
representation neurons in layer $l$ is given below.
The process for predictive activation will cycle between feedforward and feedback updates.
Eqn.~\ref{eqDerivFFloss} will be used to derive feedforward updates and 
follows the derivation in Equation Set~\ref{eq_RB_deriveOfJ}.
Eqn.~\ref{eqDerivFBloss} will be used
to derive feedback updates.

\begin{subequations}
\begin{align}
\frac{\partial \, J^{l-1}(t)}{\partial \boldsymbol{r}^l(t)} &= -W^{l,l-1} \boldsymbol{e}^{l-1} (t)\label{eqDerivFFloss}\\
\frac{\partial \, J^{l}(t)}{\partial \boldsymbol{r}^l(t)} &= \boldsymbol{r}^l(t) - \boldsymbol{\hat{r}}^l(t)=\boldsymbol{r}^l(t) - W^{l+1,l} \boldsymbol{r}^{l+1}(t)\label{eqDerivFBloss}
\end{align}
\end{subequations}

\noindent
The derivatives above give us a means to perform the predictive
activation update step (Step~2) in Free-Energy Algorithm~\ref{freeEngControl}.
They are used in the gradient-descent-based feedforward and feedback update equations given below
where we have substituted the derivatives and simplified.

\begin{subequations}
\label{eqsetRecurrentWen}
\begin{align}
\mathbf{r}^l(t+1) &= \boldsymbol{r}^l(t) - k_1 \frac{\partial \, J^{l-1}(t)}{\partial \mathbf{r}^l(t)}
                   = \boldsymbol{r}^l(t) + k_1 W^{l,l-1} \boldsymbol{e}^{l-1} (t)\label{eqFFrTplus1}\\
\mathbf{r}^l(t+1) &= \boldsymbol{r}^l(t) - \beta \frac{\partial \, J^{l}(t)}{\partial \boldsymbol{r}^l(t)}
                   = (1 - \beta) \boldsymbol{r}^l(t) + \beta (W^{l+1,l})^T \boldsymbol{r}^{l+1}(t)\label{FBrTplus1}
\end{align}
\end{subequations}

\noindent
Eqn.~\ref{eqFFrTplus1} is used for the feedforward update and 
it matches Eqn.~\ref{eq_RB_predUpdateWithT} and uses the same rate parameter $k_1$.
Eqn.~\ref{FBrTplus1} is used for the
feedback update in the recurrent cycle and does not
have an equivalent in the \cite{rao1999predictive} model.

These equations can also be compared with the free-energy model in Section~\ref{section_Friston_model}.
They have a similar purpose to those in Eqn.~\ref{eqn_bogacz_rep} and serve
as an example of implementing a free-energy-like network within a convolutional deep learning framework.
Updating this recurrent set of equations corresponds to implementing the equivalent of Step 2 of Free-Energy Algorithm~\ref{freeEngControl}.
In contrast to the free-energy model, 
the loss function is different and does not formally incorporate priors.
This latter point seems to be the most important difference from the free-energy model.
The PCN model generates predictions from the next higher layer to the preceding layer.
This is intuitive and natural,
but these predictions come from a convolutional network which does not formally incorporate
prior probabilities.
Although the model minimizes prediction errors, the prediction errors do not 
appear to be based on prior probabilities.

The data flow diagram in Fig.~\ref{purdue_group_dataflow} shows the information flow for the predictive update
step for layer $l$ in the \cite{Wen2018a} model.
The shown connectivity is based on Equations Sets~\ref{eqn_Purdue_error} and~\ref{eqsetRecurrentWen}.
When compared with Figs.~\ref{fig_predictiveElement}(b) and~\ref{fig_RB_protocal_Bogacz}(b), we see that it does not use the Rao-Ballard protocol.
Specifically, its use of top-down information is different and this
indicates that it treats prior information differently.



\subsubsection{The global prediction error learning algorithm}
We now describe how the above equations are incorporated into the predictive update cycle.
The steps below present the PCN algorithm for one batch of training.
Learning occurs only in Step~3 and this step uses a standard, backpropagation based algorithm.
The similarities of this PCN algorithm to
the Free-Energy Algorithm~\ref{freeEngControl} in Section~\ref{section_Friston_model} are
fairly clear.
Also, this algorithm is similar to the \cite{Lotter2017} algorithm with the exception
that $T$ is constrained to be one.

[Global Version]\label{algPCNglobalMain}
Operates on one input batch, $x$, of a convolutional network.

\noindent
\begin{enumerate}
\item Perform feedforward convolutional sweep on $x$.
\item Do recurrent activation updates, both feedforward and feedback, over all layers for $T$ ($\le 6$) timesteps.
      The intention is to use updates to converge the network activations to a minimum prediction error.
      An expanded description is given below.
\item Update all weights, including feedback weights, according to cross-entropy classification loss.
\end{enumerate}

Informally, the feedforward sweep in Step~1 initializes the activations for the entire network.
The predictive update in Step~2 refines these activation levels.
The feedforward weights are shared between Steps~1 and~2.
The feedback weights are only used in Step~2.
Updates in Step~2 are determined by gradient descent on the predictive error.
Following these two steps, 
Step~3 occurs where all weights (feedforward and feedback) are trained by gradient descent on the cross-entropy classification loss.
Since the recurrent network is unrolled, the weight updates affect both the feedfoward
and ultimately the recurrent weights.
Note that the classification error (cross entropy), and not prediction error, controlls the weight updates.
On the subsequent batch, the new feedforward sweep initializes the activations to improved
values as a result of the training on the previous cycle.

Steps 1 and 3, without Step 2 (or $T=0$), in PCN Algorithm~\ref{algPCNglobalMain} above is a plain feedforward algorithm
with gradient-based learning on
a convolution network for one input batch.
The novelty of global PCN lies in Step 2 where the recurrent updates take place.
To see why the above is a global algorithm,
we examine the details of the recurrent updates in Step~2.
The algorithm below expands Step~2 and is performed for $T$ iterations where $T$ is a small integer-valued hyperparameter.

\vspace{12pt}
\noindent
\textbf{Expansion of Step 2 in PCN Alg 1:}
\textit{Performs predictive activation update for one time step.}

\begin{enumerate}
\item \textit{Top-down (feedback) sweep from Layer $L$ to Layer $2$.}
     \begin{enumerate}
     \item \textit{Apply \texttt{Conv2DTranspose} to layer $l$ to generate prediction for layer $l-1$.
           Denote it by $\boldsymbol{\hat{r}}^{l-1}$ (Eqn.~\ref{eqPredictionPurdue}).}
     \item \textit{Update activations in layer $l-1$ according to a convolutional version of Eqn.~\ref{FBrTplus1}.}
     \end{enumerate} 
\item \textit{Bottom-up (feedforward) sweep from Layer $2$ to Layer $L$.}
     \begin{enumerate}
     \item \textit{Compute prediction error for Layer $l-1$ by subtracting $\boldsymbol{\hat{r}}^{l-1}$ 
           from the activations, $\boldsymbol{r}^{l-1}$ in 
           Layer $l-1$. Call this $\boldsymbol{e}^{l-1}$. (Eqn.~\ref{eqPredErrPurdue})}
     \item \textit{Update activations for Layer $l$ according to a convolutional version of Eqn.~\ref{eqFFrTplus1}.}
     \end{enumerate}
\end{enumerate}


\noindent
This is considered global update because each loop in Steps~1 and~2 traverse all of the layers
in the network and can propagate the effects of any layer to an arbitrary number of layers forward
or backward.

To summarize,
the top-down component (Step 1) computes predictions for each layer (except the top layer) and then
updates the activations of those layers to take into account the predictions.
Since the predictions for layer $l-n$ will ultimately depend on the updated activations for Layer $l$ via
Layers $l-1$, $l-2$, \ldots,
this is a global algorithm.
Updates for a higher layer can in principle influence updates for a lower layer a long distance below.
A similar long-distance effect can occur in the activation updates from the bottom-up
sweep.
For these reasons, the authors describe the predictive update algorithm as global.

\subsubsection{Local prediction error model}
\label{sec_local_prediction_error_model}
\cite{Han2018a} replaced the global predictive update scheme 
described above with a local scheme
that prevented predictive update effects from propagating to earlier layers in the network.
The modified network's classification performance did not suffer from this restriction, suggesting that
the original long-distance, top down information flow had a negligible effect on the original performance.
There are some caveats in drawing this conclusion, however, because
other aspects of the architecture were changed from the global model.
The original VGG-like architecture was replaced with
a ResNet-like architecture having skip connections.
The ResNet architecture may have
compensated for removing the global aspect of the predictive update.
Also, this local architecture used batch normalization, which is
known to improve performance, whereas 
global average pooling was used in
the previous architecture.


The local model's design has much in common with the global model.
The architecture is a feedforward convolutional network extended with recurrency
and it uses the same predictive update equations (Equation Set~\ref{eqsetRecurrentWen}).
As in the global model,
the recurrent steps are unfolded in time to transform the network into a feedforward achitecture.
The full feedforward sweep can be broken down as: 1) the classic feedforward
sweep and 2) the unrolled recurrent sweep (which becomes feedforward).
The predictive activation equations are the same as in the global model,
namely Equation Set~\ref{eqsetRecurrentWen}.


\paragraph{Training the local model}
To see how the local model differs from the global model, we examine the
network activation and training algorithm more closely.
A bottom-up sweep performs the recurrent updates between the lowest pair of consecutive layers for
$T$ times steps, and then repeats the process for the next higher layer.
Unlike PCN Algorithm~\ref{algPCNglobalMain}, there is no top-down sweep.
The algorithm is shown below.
Weight-update learning occurs in Step 2\@.

[Local Version]\label{algPCNlocalMain}
Operates on one input batch, $x$, of a convolutional network.
\begin{enumerate}
\item Perform bottom-up forward sweep. For each layer and starting with Layer $l=2$ 
      up to layer $L$ do:
     \begin{enumerate}
     \item Layer $l =$ relu(conv2D(batchNormalize layer $l-1$))
     \item Do recurrent activation updates for $T$ time steps on Layers $l$ and $l-1$ to 
           compute the converged activations of Layer $l$.
     \item Add a skip connection from Layer $l-1$ to $l$ via a bypass convolution.
     \end{enumerate}
\item Update all weights according to cross-entropy classification loss.
\end{enumerate}

\noindent
The bottom-up sweep constrains the
effect of the predictive activation updates
to affect activations only at a higher layer (and not lower layers).
The details of the recurrent updates in Step 1b are given below.

\vspace{12pt}
\noindent
\textbf{Expansion of Step 1b in PCN Alg 2:}
\textit{Performs predictive activation update for one local recurrent time step
on Layers $l$ and $l-1$.}

\begin{enumerate}
\item Apply \texttt{Conv2DTranspose} to Layer $l$ to generate prediction for Layer $l-1$.
           Denote it by $\boldsymbol{\hat{r}}^{l-1}$.
\item Compute prediction error for Layer $l-1$ by subtracting $\boldsymbol{\hat{r}}^{l-1}$ from the activations,
      $\boldsymbol{r}^{l-1}$. Denote the prediction error by $\boldsymbol{e}^{l-1}$.
\item Add $\mathtt{conv2D}(\boldsymbol{e}^{l-1})$ to the activations in Layer $l$.
\end{enumerate}

\noindent
In the above, the \texttt{conv2D} in Step~3 uses the same weights as
Step~1(a) of the overview.
There is a difference because in Step~1(a) the input is the activations of
Layer $l-1$ and the input for Step~3 is the prediction error.
That is, the same weights are used for two different purposes.

\subsection{Summary of predictive update with convolutional networks}
The previous section reviewed two papers, known as PCN models, that introduced predictive update into
deep convolutional architectures.
The predictive update served to fine tune the network activations 
to reduce prediction error after the initial feedfoward
sweep.
Gradient descent on the classification loss (cross entropy) was used to adjust the weights
after the fine tuning.
For the model that used global predictive update, 
PCN Algorithm~\ref{algPCNglobalMain}, the steps in the algorithm
are quite similar to those in Free-Energy Algorithm~\ref{freeEngControl}.
Both algorithms perform predictive update in Step~2 and weight training in Step~3.

In the PCN models, predictive update seems to prevent information loss from the representations in
the lower layers to the higher layers.
By improving the ability of the higher layer to predict the representation activations
in the lower layer, the updated representation in the higher layer 
improves information preservation of the information contained
in the lower layer.
In a plain convolutional network without predictive activation, the only factor
influencing weight update is the cross-entropy loss function designed to maximize
classification accuracy.
This goal of learning only what is needed for the immediate
classification event could filter information deemed irrelevant to the classification task, 
yet may be still relevant to longer term learning.



\paragraph{Delete?} 

A deep network 
using shared weights across layers can
be viewed as a shallow recurrent network \cite{Poggio2016a}.


\paragraph{How many training cycles did it take to converge?}
For the local version, 300 epochs with batch size of 128, and initial
learning rate of $.01$.

\textbf{Compare the trained feedfoward weights between the normal and enhanced architectures.}
The bypass connection in \cite{Han2018a} adds the representation in layer $l-1$ to 
that of layer $l$.

\textbf{Mutual information.} 
Measure the mutual information between adjacent layers before
and after predictive update.
This is a way to test whether predictive update reduces information loss.
\section{Design for revised Lotter et al.\ model using RB protocol}
\begin{figure} \centering
\includegraphics[scale=0.55]{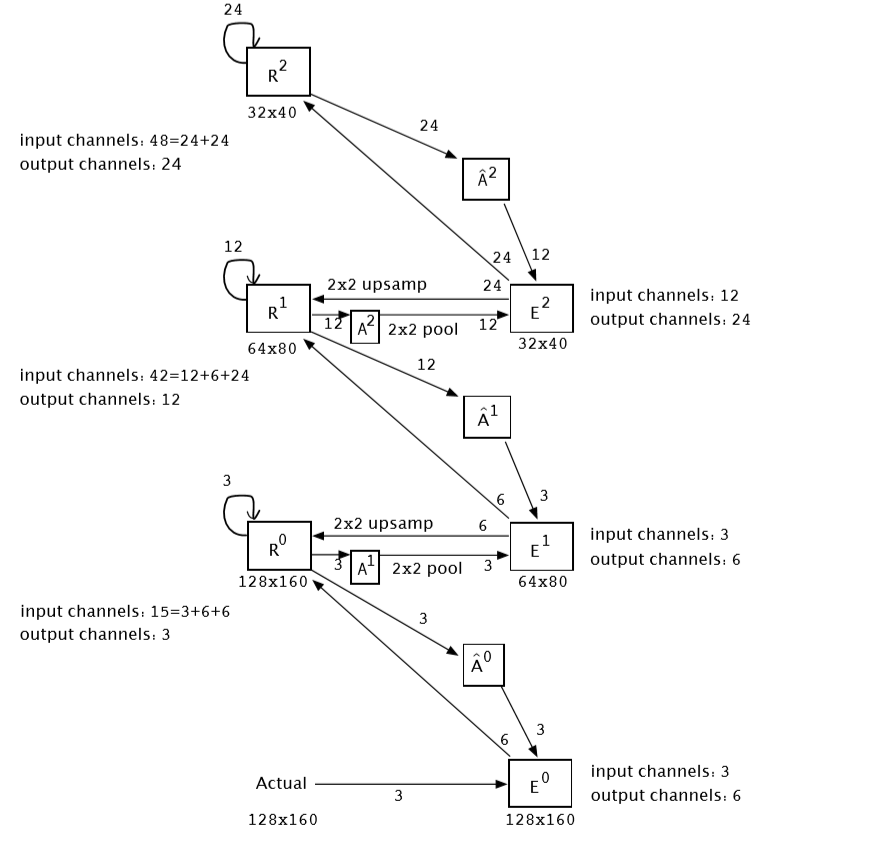}
\caption{\small Example dimensions of our 3-layer RBP architecture used in our experiments.
Labels near arrows are channel counts.
This architecture is implemented by models RB1 and RB2 in Table~\ref{tab_ModelIDsandArchitectures}.}
\label{fig_revisedLotterModelCompatibleWithRao_fleshedOut2}
\end{figure}

This appendix specifies more details of the RBP-PredNet model 
and initial experiments comparing it to 
the original PredNet \cite{Lotter2017}.
Although
our RBP model uses the same modules as the original PredNet,
their communication is reconfigured to match the RB protocol.
Fig.~\ref{fig_revisedLotterModelCompatibleWithRao_fleshedOut2} 
shows design details for a
3-layer version of our model.
In addition to the module input connectivity,
the figure shows the number of input and output channels for 
each module. 
This provides enough information to calculate the parameter count as seen
in Table~\ref{tab_RBP_param_count}.

All trainable parameters are
in the $A^l$, $\hat{A}^l$, and $R^l$ modules.
All three module 
types perform multi-channel, 2D convolutional operations.
The $A^l$ and $\hat{A}^l$ modules use one operation, whereas
the $R^l$ module, because it is a convolutional LSTM, uses four sets of identical operations.
If the number of output channels is $\mathrm{oc}$, then $\mathrm{oc}$ multi-channel convolutions
are required to calculate this output, and this is the size of the convolution set.
The convolutional LSTM has three gate operations and
one input update operation, each of which calculates one multi-channel convolution set.\footnote{Recurrent
and nonrecurrent inputs are stacked, and this contributes to the number of input channels.}
These sets are identical, except for the weight values in their kernels.
The number of input channels, denoted $\mathrm{ic}$, to an $R^l$ module is the sum of
the feedforward, lateral, and feedback inputs.
All convolutional operations use square filters with kernel size on one dimension 
denoted by $k=3$.
Taking these factors into account, 
the formula below gives the parameter count for a set of multi-channel convolutions,
henceforth called a convolution set.

\begin{equation}
\mathrm{count} = \mathrm{oc}\cdot\left(k^2\cdot\mathrm{ic}+1\right)
\end{equation}

\noindent
Above, the number of weights for a multi-channel, convolutional filter is given in parentheses.
Each filter has one bias.
For each output channel, one multi-channel convolution is needed.
Table~\ref{tab_RBP_param_count} illustrates the calculation 
for the Fig.~\ref{fig_revisedLotterModelCompatibleWithRao_fleshedOut2} 
model and reveals that the model has 65,799 trainable weights.

\begin{figure} \centering
\includegraphics[scale=0.8]{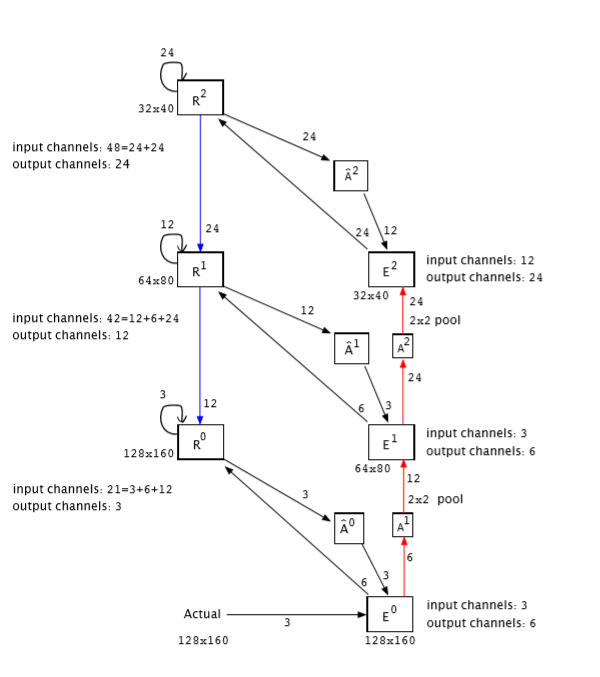}
\caption{\small Dimensions of 
a scaled down version of the original Lotter et al.\ architecture using three layers
used in our experiments. 
The blue pathway is absent from the RBP model but is used in the hybrid model.
The red pathway is missing from both the RBP and hybrid models, so is unique to the \cite{Lotter2017}
architecture.
Labels near arrows are channel counts.
This architecture is implemented by models Pred1 and Pred2 in Table~\ref{tab_ModelIDsandArchitectures}.}
\label{fig_Lotter_original_design_3_layer}
\end{figure}

We will compare the 
Fig.~\ref{fig_revisedLotterModelCompatibleWithRao_fleshedOut2} 
RBP model with a 3-layer version of the original Lotter et al.\ design
that is shown in Fig.~\ref{fig_Lotter_original_design_3_layer}.
Both variants use the same eleven modules.
Both models are constrained to use the same output channels in the $R^l$ modules.
However, to make the modules fit together,
this entails that the input channel counts for the $R^l$ modules are different,
as well as the channel counts for the $E^1$ and $E^2$ modules, and the $A^l$ and $\hat{A}^l$ modules
bridging them.
Since the input channel counts are different, 
the Fig.~\ref{fig_Lotter_original_design_3_layer} model has 
103,020 parameters (cf., Table~\ref{tab_Lotter_param_count}), instead of 65,799.


The arrows
connecting the modules in both figures show direction of information flow.
Numeric labels on the arrows 
show channel counts for
that path.
The number of output channels in the $R^l$ modules for each model
are matched.
However, the output channel count in the $E$ modules is less for the RBP model.
The image dimensions for the three layers also match.
There are more parameters in the Lotter et al.\ model than in the RBP model.
This is because the number of parameters is primarily controlled by the number of
input channels to the $R$ modules.

In both models,
the $R^0$ module has three channels because of the RGB image encoding.
The output channels
for the $E^l$ modules separately encode positive prediction error and negative prediction error,
so the number of output channels is twice the number of input channels.

\begin{table}
\begin{center}
\begin{tabular}{|l|ccccrr|}\hline
\multicolumn{7}{|c|}{Parameter Count for RBP Model}\\\hline
\multicolumn{1}{|c|}{~}&
\multicolumn{1}{c|}{\# of conv}&
\multicolumn{1}{c|}{filter}&
\multicolumn{1}{c|}{input chans}&
\multicolumn{1}{c|}{output chans}&
\multicolumn{1}{c|}{Calculation}&
\multicolumn{1}{c|}{parameter}\\
\multicolumn{1}{|c|}{Module}&
\multicolumn{1}{c|}{sets $\mathrm{(cs)}$}&
\multicolumn{1}{c|}{$k^2$}&
\multicolumn{1}{c|}{($\mathrm{ic}$)}&
\multicolumn{1}{c|}{($\mathrm{oc}$)}&
\multicolumn{1}{c|}{$\mathrm{cs}\cdot\mathrm{oc}\left(k^2\mathrm{ic}+1\right)$}&
\multicolumn{1}{c|}{count}\\\hline
$\hat{A}^0$&1&$3\times3$& 3& 3&$1\cdot3\cdot\left(3^2\cdot3 + 1\right)$                 &   84\\
$R^0$      &4 &$3\times3$&15& 3&$4\cdot3\cdot\left(3^2\cdot15 + 1\right)$&1,632\\
$A^1$      &1&$3\times3$& 3& 3&$1\cdot3\cdot\left(3^2\cdot3 + 1\right)$                 &   84\\
$\hat{A}^1$&1&$3\times3$&12& 3&$1\cdot3\cdot\left(3^2\cdot12+1\right)$        &  327\\
$R^1$      &4&$3\times3$&42&12&$4\cdot12\cdot\left(3^2\cdot42+1\right)$ &18,192\\
$A^2$      &1&$3\times3$&12&12&$12\cdot\left(3^2\cdot12+1\right)$       & 1,308\\
$\hat{A}^2$&1&$3\times3$&24&12&$12\cdot\left(3^2\cdot24+1\right)$       & 2,604\\
$R^2$      &4&$3\times3$&48&24&$4\cdot24\cdot\left(3^2\cdot48+1\right)$ &41,568\\\hline
~          &~&~         &~ &~ &Total parameters:                            &65,799\\\hline
\end{tabular}
\end{center}
\caption{Parameter count for the RBP model depicted in Fig.~\ref{fig_revisedLotterModelCompatibleWithRao_fleshedOut2}.
If the LSTMs in the $R^l$ models are replaced by GRUs, then the parameter count is 50,451 instead of 65,799.
This is obtained by changing the number of convolution sets from four to three in each of the $R^l$ modules.}
\label{tab_RBP_param_count}
\end{table}

\begin{figure} \centering
\includegraphics[scale=0.8]{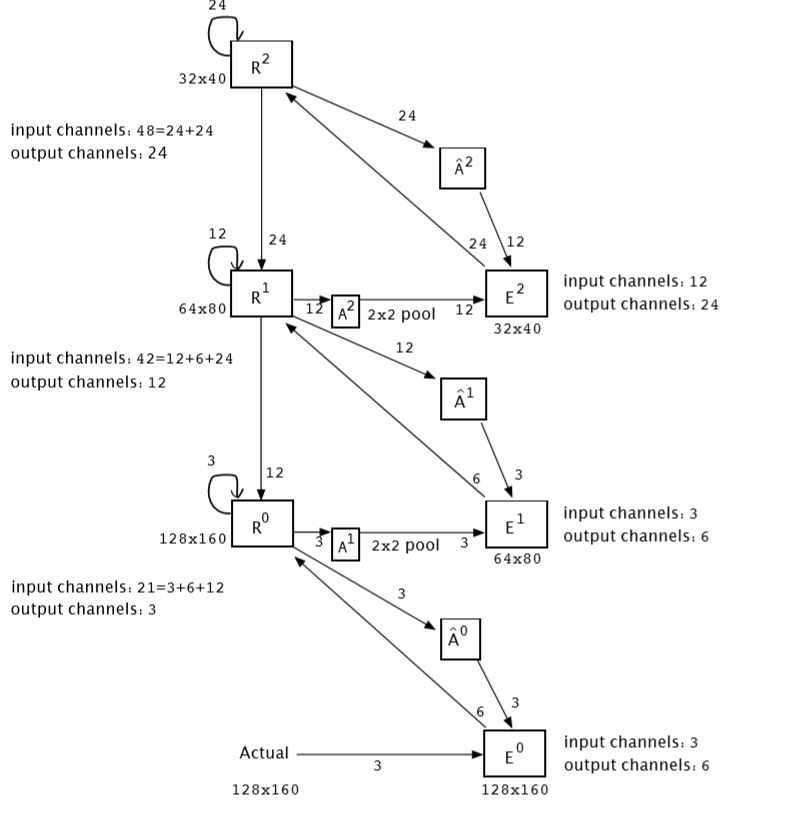}
\caption{\small Example dimensions of our 3-layer hybrid architecture used in our experiments.
Labels near arrows are channel counts.}
\label{fig_revisedLotterModel_hybridV3}
\end{figure}

\begin{table}
\begin{center}
\begin{tabular}{|l|ccccrr|}\hline
\multicolumn{7}{|c|}{Parameter Count for Lotter et al.\ PredNet Model}\\\hline
\multicolumn{1}{|c|}{~}&
\multicolumn{1}{c|}{\# of conv}&
\multicolumn{1}{c|}{filter}&
\multicolumn{1}{c|}{input chans}&
\multicolumn{1}{c|}{output chans}&
\multicolumn{1}{c|}{Calculation}&
\multicolumn{1}{c|}{parameter}\\
\multicolumn{1}{|c|}{Module}&
\multicolumn{1}{c|}{sets $(\mathrm{cs})$}&
\multicolumn{1}{c|}{$k^2$}&
\multicolumn{1}{c|}{($\mathrm{ic}$)}&
\multicolumn{1}{c|}{($\mathrm{oc}$)}&
\multicolumn{1}{c|}{$\mathrm{cs}\cdot\mathrm{oc}\left(k^2\mathrm{ic}+1\right)$}&
\multicolumn{1}{c|}{count}\\\hline
$\hat{A}^0$&1&$3\times3$& 3& 3&$1\cdot3\cdot\left(3^2\cdot3 + 1\right)$                 &     84\\
$R^0$      &4&$3\times3$&21& 3&$4\cdot3\cdot\left(3^2\cdot21 + 1\right)$                &  2,280\\
$A^0$      &1&$3\times3$& 6&12&$1\cdot12\cdot\left(3^2\cdot6 + 1\right)$                &    660\\
$\hat{A}^1$&1&$3\times3$&12&12&$1\cdot12\cdot\left(3^2\cdot12+1\right)$                 &  1,308\\
$R^1$      &4&$3\times3$&60&12&$4\cdot12\cdot\left(3^2\cdot60+1\right)$                 & 25,968\\
$A^1$      &1&$3\times3$&24&24&$1\cdot24\cdot\left(3^2\cdot24+1\right)$                 &  5,208\\
$\hat{A}^2$&1&$3\times3$&24&24&$1\cdot24\cdot\left(3^2\cdot24+1\right)$                 &  5,208\\
$R^2$      &4&$3\times3$&72&24&$4\cdot24\cdot\left(3^2\cdot72+1\right)$                 & 62,304\\\hline
~          &~&~         &~ &~ &Total parameters:                                        &103,020\\\hline
\end{tabular}
\end{center}
\caption{Parameter count for the Lotter et al.\ model shown in Fig.~\ref{fig_Lotter_original_design_3_layer}.}
\label{tab_Lotter_param_count}
\end{table}

\textbf{Method.}
All models were trained
for twenty epochs, using the Adam optimizer,
on the preprocessed KITTI traffic data set used
in \cite{Lotter2017}.
This data set was preprocessed by \cite{Lotter2017}
to obtain three-channel color images of size $120 \times 160$ pixels.
Successful prediction on this dataset requires the model to detect and track
several moving
and nonmoving objects in a video frame.

Three architectures were tested.
The first was the RBP architecture where the $R^l$ modules were built from
convolutional LSTMs. 
The second was the RBP architecture where the $R_l$ modules were built from
convolutional GRUs.
The third was tested using the original \cite{Lotter2017} architecture.
Two sets of loss function weights were used,
namely, $[.5,.4, .2]$ and $[1, 0, 0]$.

Several architectures were tested as specified
in Table~\ref{tab_ModelIDsandArchitectures}.

The first experiment used the loss function weights with values $[.5,.4, .2]$.
The second experiment used loss function weights with values $[1, 0, 0]$.
Two experiments were performed.
We also studied a GRU based version of the RBP model.
This replaced the LSTMs in the $R^l$ modules with GRUs.

For all experiments,
three performance measures were recorded:
mean absolute error (MAE),
mean squared error (MSE), and structural similarity index (SSIM).
All measures were calculated
on a baseline control condition \cite{Lotter2017} for comparison with
the neural network performance measures.
The baseline used the current video frame as the prediction for the
next frame.

\begin{table}
\begin{center}
\begin{tabular}{|l|rrrrr|}\hline
\multicolumn{6}{|c|}{Model Parameters as a Function of ID}\\\hline
\multicolumn{1}{|c|}{Model ID}&
\multicolumn{1}{|c|}{R type}&
\multicolumn{1}{c|}{stack\_sizes}&
\multicolumn{1}{c|}{R\_stack\_sizes}&
\multicolumn{1}{c|}{Loss weights}&
\multicolumn{1}{c|}{Params}\\\hline\hline
Pred1   &LSTM&[3, 12, 24]&[3, 12, 24]&[.5, .4, .2]&103,020\\\hline
Pred2   &LSTM&[3, 12, 24]&[3, 12, 24]&[1.0, 0, 0] &103,020\\\hline
RB1     &LSTM&[3, 3, 12] &[3, 12, 24]&[.5, .4, .2]&65,799\\\hline
RB2     &LSTM&[3, 3, 12] &[3, 12, 24]&[1.0, 0, 0] &65,799\\\hline
RB3     &LSTM&[3, 12, 24]&[10, 16, 30]&[1.0, 0, 0] &162,641\\\hline
RB3\_gru&GRU &[3, 12, 24]&[10, 16, 30]&[.5, .4, .2] &125,523\\\hline
RB4\_gru&GRU &[3, 12, 24]&[10, 16, 30]&[1.0, 0, 0] &125,523\\\hline
RB5     &LSTM&[3, 3 , 12]&[ 3, 12, 24]&[.33, .33, .33]&65,799\\\hline
RB6     &LSTM&[3, 3]     &[ 3, 12]    &[1.0, 0]       & 9,951\\\hline
RB7     &LSTM&[3, 3]     &[ 3, 12]    &[.5, .5]       & 9,951\\\hline
\end{tabular}
\end{center}
\caption{Model architecture specifications indexed under Model ID.
``stack\_sizes'' is the number of input channels to the error modules.
``R\_stack\_sizes'' is the number of output chanels of the representation modules.
``Params'' is the number of trainable parameters in the model.}
\label{tab_ModelIDsandArchitectures}
\end{table}

\begin{table}
\begin{center}
\begin{tabular}{|l|lccccc|}\hline
\multicolumn{1}{|c|}{Model}&
\multicolumn{1}{|c|}{Model}&
\multicolumn{1}{c|}{MAE}&
\multicolumn{1}{c|}{MAE}&
\multicolumn{1}{c|}{MAE}&
\multicolumn{1}{c|}{MSE}&
\multicolumn{1}{c|}{SSIM}\\
\multicolumn{1}{|c|}{Type}&
\multicolumn{1}{|c|}{ID}&
\multicolumn{1}{c|}{train}&
\multicolumn{1}{c|}{valid}&
\multicolumn{1}{c|}{predict}&
\multicolumn{1}{c|}{predict}&
\multicolumn{1}{c|}{predict}\\\hline\hline
Baseline&~    &~    &~    &.0757&.0212&.6750\\
Lotter  &Pred1&.0187&.0243&.0694&.0162&.7138\\
RBP     &RB1  &.0191&.0245&.0704&.0163&.7116\\
RBP\_gru&RB3\_gru&.0195&.0245&\textbf{.0692}&\textbf{.0160}&\textbf{.7159}\\
\hline
\end{tabular}
\end{center}
\caption{Comparisons for the original Lotter model with LSTM and the RBP model using LSTM and GRU\@.
Layer loss function weights for both models was $[.5, .4, .2]$.
The full model specification
can be found by looking up Model~ID in
Table~\ref{tab_ModelIDsandArchitectures}.
(MAE: mean absolute error.
MSE: mean squared error.)}
\label{tab_firstResults_comparing_RPB_performance_with_Lotter}
\end{table}

\textbf{Results.}
The performance results appear in Tables~\ref{tab_firstResults_comparing_RPB_performance_with_Lotter} 
and~\ref{tab_firstResults_comparing_RPB_performance_with_Lotter2}.
The simulations in Table~\ref{tab_firstResults_comparing_RPB_performance_with_Lotter}
use loss function weights of $[.5,.4, .2]$
and the simulations in Table~\ref{tab_firstResults_comparing_RPB_performance_with_Lotter2} use loss 
function weights of $[1, 0, 0]$.
The prediction error scores should be compared with the appropriate baseline scores.

The results for the first experiment appear in
Table~\ref{tab_firstResults_comparing_RPB_performance_with_Lotter} and in
Fig.~\ref{fig_comparing_RBP_performance_with_Lotter1}(a) and (b).
Given the Model ID, the architecture can be found in Table~\ref{tab_ModelIDsandArchitectures}.
During training, the RBP model achieves a mean absolute error (MAE) for training data
of $.0191$ and for validation error it achieves an MAE of $.0245$.
For testing, the MSE next frame prediction accuracy is $.0163$ in comparison to a
baseline prediction accuracy of $.0212$ that uses the previous frame as the prediction.
These results are very close to the comparable \cite{Lotter2017} model.
Fig.~\ref{fig_comparing_RBP_performance_with_Lotter1}(a) and (b) shows that the qualitative 
structure of the training loss curves are virtually
identical.
Tables~\ref{tab_firstResults_comparing_RPB_performance_with_Lotter} 
and~\ref{tab_firstResults_comparing_RPB_performance_with_Lotter2} 
also so the structural similarity index (SSIM).
This gives a better measure of perceived image similarity than MSE \cite{Wang2004a}.
These results indicate that the two models are virtually equivalent despite their different
communication structures.

The second experiment with loss weights of $[1.0, 0., 0.]$ showed a different pattern.
The prediction performance of both models improved.
In the case of the Lotter et al.\ version of the model, this is expected because 
the result was reported
in the original paper. \cite{Lotter2017}.

\begin{figure} \centering

\subfloat[]
{
\includegraphics[scale=0.5]{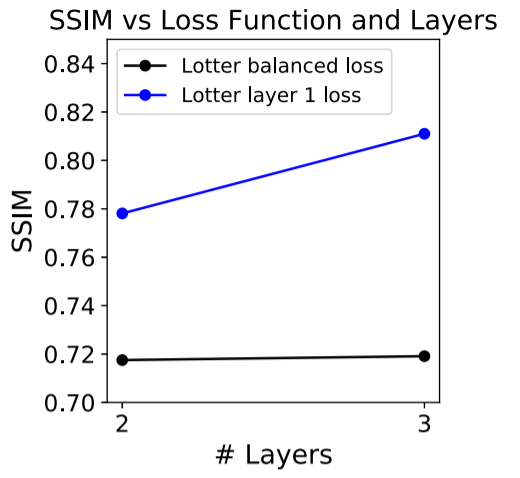}
}
\subfloat[]
{
\includegraphics[scale=0.35]{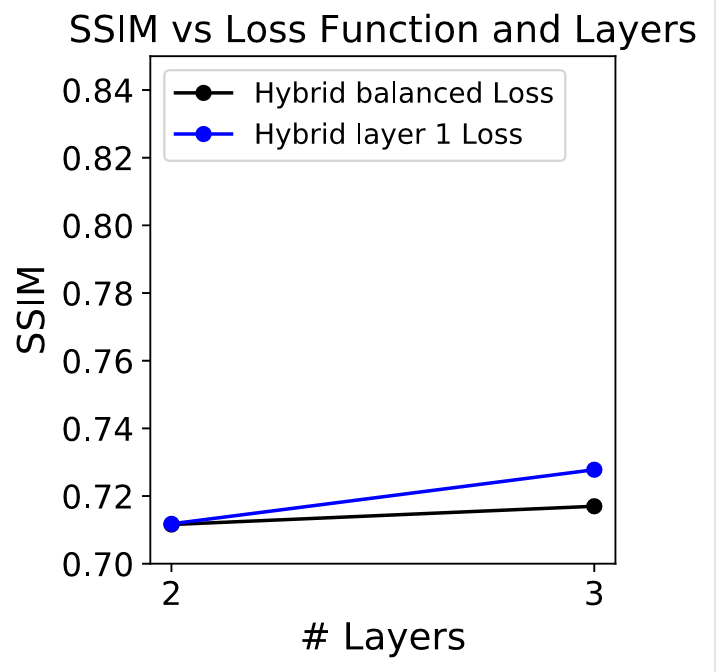}
}

\caption{\small SSIM performance of the Lotter architecture (lefft) and for the hybrid architecture (right).
Number of layers versus type of loss weight set was compared.
Trained for 50 epochs.}
\label{fig_hybridV3_losswts_vs_layers}
\end{figure}

\begin{figure} \centering

\subfloat[]
{
\includegraphics[viewport=-0.2in 0.0in 5.5in 3.8in,clip=true,scale=.45]{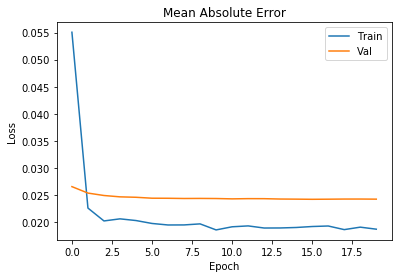}
} 
\subfloat[]
{
\includegraphics[viewport=-0.45in 0.0in 7.0in 3.8in,clip=true,scale=.45]{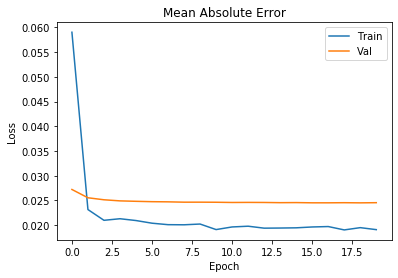}
}
\caption{Training and validation performance as measured by mean absolute error
and loss = $[.5,.4,.2]$.
(a) Original \cite{Lotter2017} version of PredNet.
(b) Novel RBP version of PredNet.}
\label{fig_comparing_RBP_performance_with_Lotter1}
\end{figure}

\textbf{Testing the hybrid model.}
Four experiments were performed to test the hybrid model.
We assessed SSIM performance on testing data for two- and three-layer versions of the model,
after for training for 50 epochs.
We also tested two types of
loss weights.
These were first balanced, [.5, .5] and [.33, .33, .33], for the two- and three-layer models, respectively.
Secondly, we tested loss weights that used the first layer only, that is [1, 0] and [1, 0, 0].
The results are shown in Figure~\ref{fig_hybridV3_losswts_vs_layers}.

From the figure, we see that the two-layer SSIM performance was the same for both types of
loss function.
Also, regardless of loss function, three-layer performance was better than two-layer performance.
Finally, the layer-1-only loss improved performance for the three layer network.

\textbf{Discussion.}
One would expect that loss weights balanced across the layers (or weighted by precision)
would yield better performance in a predictive
coding model.
Empirically, we observed that it is not true.
Shifting all of loss to the first layer gives better performance.

However, in retrospect, the above results makes sense.
The layer 1 loss weights emphasize to the task at hand.
Reducing the loss in a higher layer is irrelevant if it does not
improve performance in the target task.
Using balanced loss weights might produce a more versatile representation allowing
improved transfer learning.


\begin{table}
\begin{center}
\begin{tabular}{|l|lccccc|}\hline
\multicolumn{1}{|c|}{Model}&
\multicolumn{1}{|c|}{Model}&
\multicolumn{1}{c|}{MAE}&
\multicolumn{1}{c|}{MAE}&
\multicolumn{1}{c|}{MAE}&
\multicolumn{1}{c|}{MSE}&
\multicolumn{1}{c|}{SSIM}\\
\multicolumn{1}{|c|}{Type}&
\multicolumn{1}{|c|}{ID}&
\multicolumn{1}{c|}{train}&
\multicolumn{1}{c|}{valid}&
\multicolumn{1}{c|}{predict}&
\multicolumn{1}{c|}{predict}&
\multicolumn{1}{c|}{predict}\\\hline\hline
Baseline&        &     &     &.0757&.0212&.6750\\
Lotter  &Pred2   &.0295&.0374&\textbf{.0568}&\textbf{.0109}&\textbf{.7906}\\
RBP     &RB2     &.0378&.0486&.0699&.0160&.7137\\
RBP     &RB3     &.0371&.0482&.0683&.0155&.7196\\
RBP\_gru&RB4\_gru&.0384&.0482&.0685&.0154&.7210\\
\hline
\end{tabular}
\end{center}
\caption{Comparisons for the original Lotter model and the RBP model for three layers.
The loss function weights for both models is $[1.0, .0, .0]$.
More information about the model can be found by looking up Model~ID in
Table~\ref{tab_ModelIDsandArchitectures}.
MAE: mean absolute error.
MSE: mean squared error.}
\label{tab_firstResults_comparing_RPB_performance_with_Lotter2}
\end{table}

\begin{table}
\begin{center}
\begin{tabular}{|l|lccccccc|}\hline
\multicolumn{1}{|c|}{Model}&
\multicolumn{1}{|c|}{Model}&
\multicolumn{1}{c|}{MAE}&
\multicolumn{1}{c|}{MAE}&
\multicolumn{1}{c|}{MAE}&
\multicolumn{1}{c|}{MSE}&
\multicolumn{1}{c|}{MSE}&
\multicolumn{1}{c|}{MSE}&
\multicolumn{1}{c|}{SSIM}\\
\multicolumn{1}{|c|}{Type}&
\multicolumn{1}{|c|}{ID}&
\multicolumn{1}{c|}{train}&
\multicolumn{1}{c|}{valid}&
\multicolumn{1}{c|}{predict}&
\multicolumn{1}{c|}{train}&
\multicolumn{1}{c|}{valid}&
\multicolumn{1}{c|}{predict}&
\multicolumn{1}{c|}{predict}\\\hline\hline
Baseline&        &     &     &.0757&&&.0212&.6750\\
RBP     &RB5     &.0126&.0162&.0705&     &     &.0163&.7116\\
RBP     &RB6     &.0386&.0488&.0701&.0017&.0026&.0161&.7134\\
RBP     &RB7     &.0193&.0245&.0710&     &     &.0164&.7107\\
\hline
\end{tabular}
\end{center}
\caption{New data for the RBP model. The main points are, first, that removing the third layer,
leading to a great reduction in parameters has very little negative effect on 
network SSIM\@. Second, changing the loss weights does not affect SSIM very much.
More information about the model can be found by looking up Model~ID in
Table~\ref{tab_ModelIDsandArchitectures}.
MAE: mean absolute error.
MSE: mean squared error.}
\label{tab_firstResults_comparing_RPB_performance_with_Lotter3}
\end{table}

\begin{figure} \centering

\subfloat[]
{
\includegraphics[viewport=-0.2in 0.0in 5.5in 3.8in,clip=true,scale=.45]{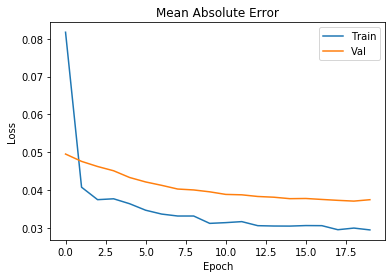}
} 
\subfloat[]
{
\includegraphics[viewport=-0.4in 0.0in 7.0in 3.8in,clip=true,scale=.45]{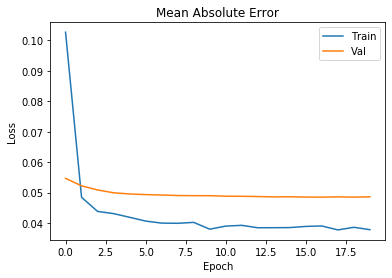}
}
\caption{Training and validation performance as measured by mean absolute error
and loss = $[1.0,0,0]$.
(a) Original \cite{Lotter2017} version of PredNet.
(b) Novel RBP version of PredNet.}
\label{fig_comparing_RBP_performance_with_Lotter2}
\end{figure}

\begin{figure} \centering

\subfloat[]
{
\includegraphics[viewport=-0.2in 0.0in 5.5in 3.8in,clip=true,scale=.45]{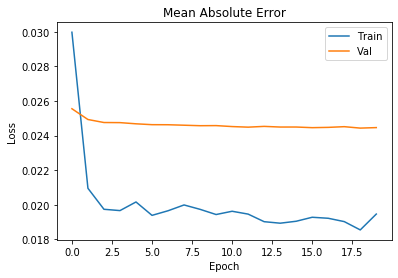}
} 
\subfloat[]
{
\includegraphics[viewport=-0.4in 0.0in 7.0in 3.8in,clip=true,scale=.45]{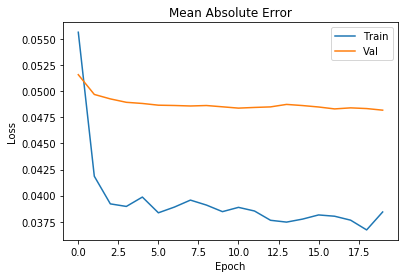}
}
\caption{Training and validation performance as measured by mean absolute error
and loss = $[1.0,0,0]$.
(a) RB3\_gru $[.5, .4, .2]$.
(b) RB4\_gru $[1.0, 0, 0]$.}
\label{fig_comparing_RBPgrus_loss_wts}
\end{figure}

\section{Summary and Conclusions}
We have presented a brief overview of some of the core predictive coding models.
It should be noted that we have omitted discussing a highly influential model \cite{Friston2005,Bogacz2017}
because it is too advanced to explain in the space available.
It is however essential reading for those interested in the subject.
This preparatory survey forms a good foundation for approaching those more
advanced topics.

Early predictive coding models \cite{rao1999predictive,Friston2005,Spratling2010a} all use the Rao-Ballard protocol as the
underpinning of their architecture.
This protocol is well grounded in Bayesian generative frameworks.
Models that deviate from this protocol (e.g., \cite{Lotter2017}) 
should justify their reasons and also explain the trade-offs they considered
when developing their model.

Earlier, we had stated that the RB protocol provides a rubric to assess the fidelity of deep learning models
that claim to implement predictive coding.
The \cite{Lotter2017} model built a hierarchy of error-driven 
representations but it deviates from the Rao-Ballard protocol.
The inputs to construct the higher-level representations were prediction errors, in 
contrast to lower level representations,
coming from the immediately preceding layer.
In an error hierarchy framework, it is not obvious
what would cause hierarchical feature representations to arise?
Predictive errors trigger learning but is there anything 
more needed to trigger learning of feature hierarchies?

What kinds of representations are created by predictive coding?
By definition, predictive coding learning algorithms 
reduce prediction error.
If these representations arise only by reducing prediction error,
then there is no a priori reason
that the representations are useful for other purposes, such as classification.
The Purdue research group, using convolutional networks for classification,
has provided results relevant to this question.

This is not the only possible type of hierarchy.
More commonly, instead of an error hierarchy,
we see a feature hierarchy typical of a convolutional network.
One expects to detect more complex, or higher-level, features at
higher layers in the hierarchy.
In a purely feedforward convolutional network, the creation of higher-level features
is an emergent property of the supervised learning
algorithm operating on the convolutional architecture.
The higher-level features apparently arise from decreasing the image size
and increasing the number of channels in the feature maps of layers
deeper in the hierarchy.

\section{Other approaches not involving predictive coding}
Although the present manuscript is about predictive coding,
we must address the question of whether predictive coding is
essential for a brain-like, intelligent system.
Thus, it is a good idea to examine other approaches which do not
use predictive coding and their reasons for success.

Architectures which have these properties are autoencoders
and deep belief networks.

\subsection{Recurrent computations and pattern completion}
Implementations of predictive update always use recurrent computation in some form.
In contrast,
\cite{Tang2018a} used recurrent computations to improve category recognition of
poorly visible and occluded
objects, but their model did not use predictive coding.
Their premise was that category
recognition of poorly visible objects is performed by a pattern completion
process and the classic, recurrent
Hopfield network \cite{Hopfield82} can serve as a model for this.
Let's compare this strategy with predictive coding.

\cite{Tang2018a} present three lines of empirical evidence consistent with the hypothesis of recurrent 
computation underlying recognition of poorly visible and partially occluded objects.
The first line concerns backward masking phenomena in humans.
Backward masking does not disrupt the
recognition of fully visible objects, but it does disrupt the recognition of 
occluded objects.
The phenomenon is interpreted is allowing the faster feedforward computations to complete while
disproportionately
preventing the slower recurrent computations from coming into play, thereby impairing the reconstruction
of occluded objects.
The feedforward computations are sufficient to recognize fully visible objects.
The second line of evidence is that
measured response latencies in the human ventral pathway to visual stimuli increase with partially visible
or occluded stimuli.
These increased latencies are consistent with the need for recurrent computations to
iteratively perform the pattern completion.
Finally,
category recognition of AlexNet \cite{Krizhevsky2012a} and other more recent state-of-the-art \cite{Simonyan2015a} 
purely feedforward artificial neural networks was shown to degrade
dramatically with occlusion and partial visibility.
All studied feedforward networks were pretrained and used five categories.
This lack of robustness is consistent with the need for
additional computations to support pattern completion of degraded objects.

\cite{Tang2018a} showed that adding Hopfield recurrency to even a single layer of a feedforward network
greatly improved the network's ability to perform pattern completion.
The recurrency took the form of a Hopfield network in which weights were preset to create
attractors that allowed a predetermined small set of
categories to be recognized via pattern completion.
Their initial tests used AlexNet and added Hopfield recurrency to layer fc7\@.
This layer had 4096 features (units), so the associated recurrent network 
was fully connected and used these features.
The weights were preset to use the five categories as attractors.
Because the original Hopfield networks had low storage capacity \cite{Hertz91},
it is an open question as to whether this approach would scale to
a much larger number of categories.
Specifically, the capacity of a binary Hopfield network that stores uncorrelated
patterns is about 0.138.
This means that a network with $N$ units has a capacity of about $0.138N$ random patterns.
Although both the feedforward weights were pretrained and the recurrent weights were 
preset, they were ultimately governed by cost functions.
For the feedforward network, the cost function reduced classification error.
For the Hopfield network, the weights were set (via the generalized Hebbian rule) so
to create attractors for categories such that there were at locally minimal energy states.
Neither of these cost functions are related to predictive coding.

\vspace{12pt}
\noindent
\textbf{Research questions related to pattern completion:}
\begin{enumerate}
\item Can an STDP-based spiking network implement a Hopfield attractor network?
\item Can we develop a predictive coding architecture for pattern completion?
      It seems to be an inherently predictive problem.
      Perhaps the Purdue group's models can work in this situation.
\end{enumerate}

\subsection{Other ways to reduce reconstruction error}
There are two other approaches to creating hierarchical representations by reducing reconstruction
error, namely, stacked auto-encoders and deep belief networks (which are stacked restricted
Boltzmann machines).
Examining them may offer insights into whether predictive
coding is essential.

\section{Open questions}
How does the residual error that guides the construction of higher-level representations
improve the learning ability of the \cite{Lotter2017}
model (How?).
Since the only purpose of the representations in the \cite{Lotter2017} model is to reduce
higher-order residual errors,
they may not form good stand alone representations for other tasks such as object classification.

\section{Predictive coding as the basis for a unified cognitive architecture}
\begin{enumerate}
\item Next-frame prediction
\item Classification
\item Anomaly detection.
There seems to be a relationship between prediction failures and anomaly (outlier) detection.
It has been hypothesized that the residual errors in the higher layers of the \cite{Lotter2017} PredNet
could be used for anomaly detection because they signal prediction errors.
See Matin's write-up. Also look at Fig 2 of 2018 Lotter paper.
\end{enumerate}

\bibliographystyle{abbrv}
\bibliography{main}

\appendix
\section{Appendix: How the Keras code for Cox Lab PredNet works}
The Keras code runs on the KITTI data set and, when used in training mode,
has two main files: \texttt{kitti\_train.py} and \texttt{prednet.py}.
To train the model, execute \texttt{kitti\_train.py} which will call \texttt{prednet.py} to build the model,
before training begins.

The file
\texttt{prednet.py} builds a custom PredNet layer. 
Within that file, the class \texttt{PredNet} is subclassed from
the abstract class \texttt{Recurrent} (which is a legacy class and is in turn
subclassed from \texttt{Layer}).
The \texttt{PredNet} class has the methods:
\begin{enumerate}
\item \texttt{\_\_\_init\_\_\_()}
      A constructor that initializes a set of instance variables associated with \texttt{self}.
      Examples are the variables \texttt{stack\_sizes} and \texttt{R\_stack\_sizes}. The former codes
      the number of channels in targets and predictions for each layer. The latter codes the number of
      channels in the representation module for each layer.
\item \texttt{compute\_output\_shape()}
      Needed for any custom layer.
      Lets Keras do automatic shape inference.
\item \texttt{get\_initial\_state()}
\item \texttt{build()} 
      This builds all of the \texttt{Conv2D} objects used in the original Prednet.
      It turns out that the modified Prednet uses exactly the same modules as the original.
      Only the information flow between the modules needs to be changed 
      to implement the Rao-Ballard protocol.
      This is done in \texttt{step()}.
\item \texttt{step()}
      The update operations take place by explicit calls to inputs for the gates.
      The inputs are created in a customized fashion.
      The purpose of this appears to be to implement the unique top-down and bottom-up sweeps on
      the representation and error modules, respectively.
\item \texttt{get\_config()}
\end{enumerate}

\noindent
The code in this file builds custom convolutional LSTMs and does not use
the Keras \texttt{convLSTM2D} API\@.
However it does import from \texttt{keras.layers} the classes: \texttt{Recurrent}, \texttt{Conv2D},
\texttt{UpSampling2D}, and \texttt{MaxPooling2D}.

The code in file \texttt{kitti\_train.py} builds and runs a multi-layer model on the KITTI data set.
When this happens, 
many of the methods in \texttt{prednet.py} are called implicitly.

\section{Appendix: How the PyTorch code for Purdue PredNet works}
The PyTorch code runs on the CIFAR and ImageNet data sets
and goes with the paper \cite{Han2018a} that describes the local model and can be found at: \texttt{github.com/libilab}.
We describe here the code that runs on the CIFAR data set.
This code has three files: \texttt{main\_cifar.py}, \texttt{prednet.py}, and \texttt{utils.py}.

\end{document}